\documentclass{article}
\usepackage{lineno,hyperref}
\modulolinenumbers[5]
\usepackage{soul,color}

\usepackage{array}
\newcolumntype{C}[1]{>{\centering\let\newline\\\arraybackslash\hspace{0pt}}m{#1}}
\newcolumntype{P}[1]{>{\centering\arraybackslash}p{#1}}
\newcolumntype{M}[1]{>{\centering\arraybackslash}m{#1}}

\usepackage{footnote}
\usepackage[symbol]{footmisc}
\makesavenoteenv{tabular}
\makesavenoteenv{table}
\usepackage{booktabs} % For formal tables
\usepackage[ruled]{algorithm2e} % For algorithms
\usepackage{txfonts}
\usepackage{graphicx}
\usepackage{subcaption}
\usepackage{lmodern}
\usepackage{multirow}

\usepackage{arxiv}

\usepackage[utf8]{inputenc} % allow utf-8 input
\usepackage[T1]{fontenc}    % use 8-bit T1 fonts
\usepackage{hyperref}       % hyperlinks
\usepackage{url}            % simple URL typesetting
\usepackage{booktabs}       % professional-quality tables
\usepackage{amsfonts}       % blackboard math symbols
\usepackage{nicefrac}       % compact symbols for 1/2, etc.
\usepackage{microtype}      % microtypography
\usepackage{lipsum}		% Can be removed after putting your text content
\usepackage{graphicx}

\title{Finger Texture Biometric Characteristic: A Survey}

	\author{Raid R. O. Al-Nima \\
	Technical Engineering College of Mosul, \\Northern Technical University, Mosul, Iraq\\
	\texttt{raidrafi3@ntu.edu.iq} \\
	\And
	Tingting Han \\
	Department of Computer Science and Information Systems, \\Birkbeck, University of London, London, UK\\
	\texttt{tingting@dcs.bbk.ac.uk} \\
	\AND
	Taolue Chen \\
	Department of Computer Science and Information Systems, \\Birkbeck, University of London, London, UK\\	
	\texttt{taolue@dcs.bbk.ac.uk} \\
	\And
	Satnam Dlay \\
	School of Engineering, Newcastle University, \\
	Newcastle upon Tyne, UK\\
	\texttt{satnam.dlay@ncl.ac.uk} \\
	\And
	Jonathon Chambers \\
	School of Engineering, University of Leicester \\
	Leicester, UK \\
	\texttt{Jonathon.Chambers@le.ac.uk} \\
}

\begin{document}
\maketitle

\begin{abstract}
	In recent years, the Finger Texture (FT) has attracted considerable attention as a biometric characteristic. It can provide efficient human recognition performance, because it has different human-specific features of apparent lines, wrinkles and ridges distributed along the inner surface of all fingers. Also, such pattern structures are reliable, unique and remain stable throughout a human's life. Efficient biometric systems can be established based only on FTs. In this paper, a comprehensive survey of the relevant FT studies is presented. We also summarise the main drawbacks and obstacles of employing the FT as a biometric characteristic, and provide useful suggestions to further improve the work on FT. 
\end{abstract}

% keywords can be removed
\keywords{Finger Texture \and Finger Inner Surface \and Biometrics \and Pattern Recognition}

\section{Introduction}
\begin{figure}[!h]
	\centering
	\includegraphics[scale=.4,trim=4cm 12.5cm 4cm 12cm,clip]{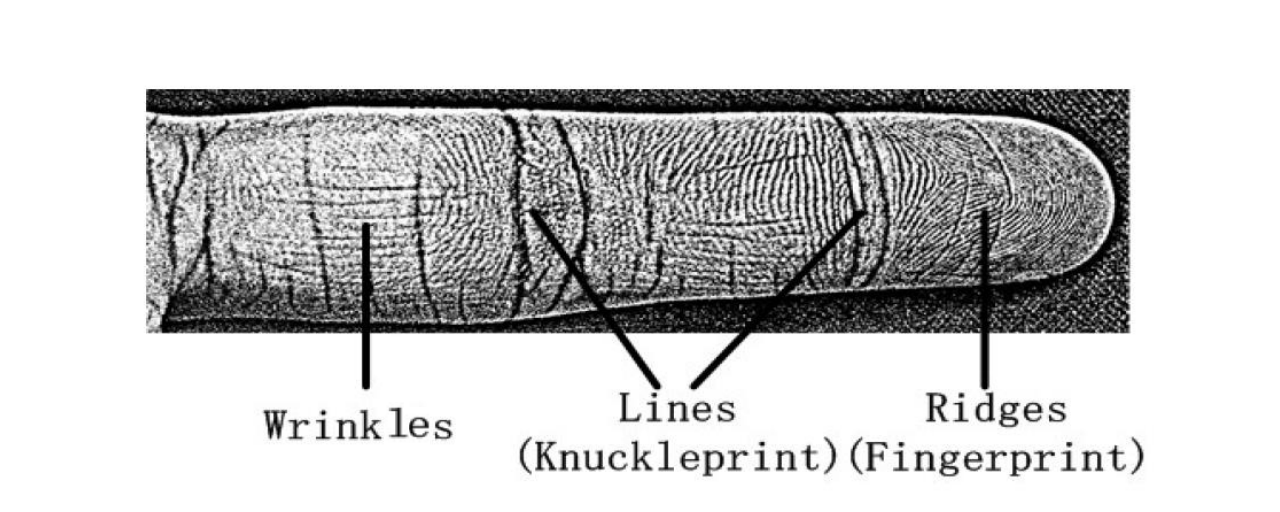}
	\caption{Various patterns that form the inner surface of a finger: ridges, visible lines and skin wrinkles as given in \cite{li2004personal}}	\label{fig:FT2}
\end{figure}

The inner surface of the finger has been the subject of several recent investigations. It has similar features as can be observed in the palm surface. The main features of the inner finger surface are shown in Fig. \ref{fig:FT2}. It includes various patterns of ridges, visible lines and skin wrinkles. In general, the ridges require higher resolution images than the lines and wrinkles. As a result, the visible patterns of the flexion lines and wrinkles becomes simpler and more effective features \cite{li2004personal} (than ridges). 

\begin{figure}[!h]
	\centering
	\includegraphics[page=1,scale=.35,trim=3cm 8cm 3cm 8cm,clip]{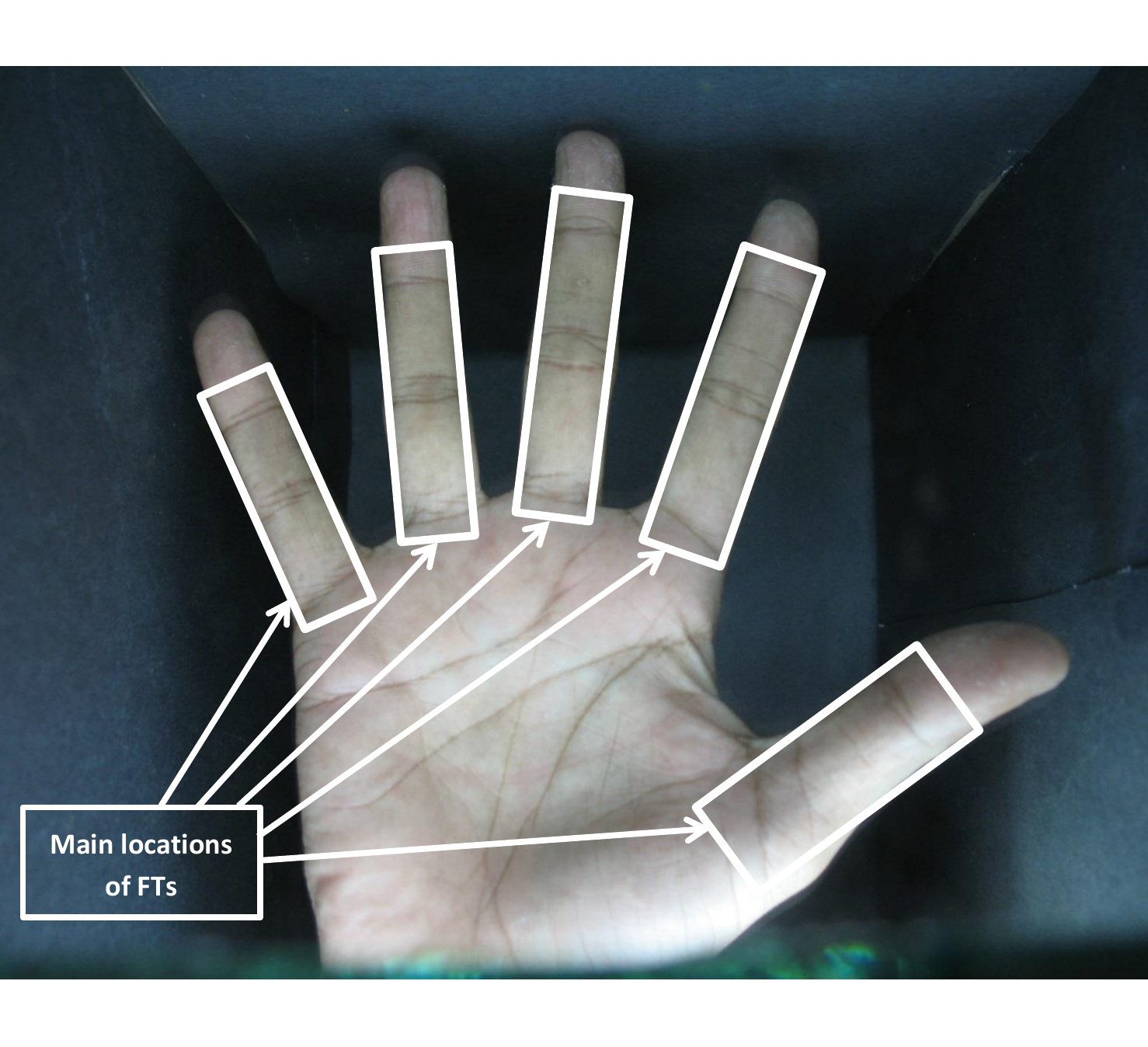}
	\caption{The main positions of the FTs (assigned by the white rectangles). Essentially, %as explained in \cite{Al-Nima2017Signal}, 
		they can be found in the inner hand surface of the five fingers and they are located between the upper phalanx (below the fingerprint) and the lower knuckle (the base of the finger)}
	\label{fig:FTs_locations}
\end{figure}

The FT is defined as the skin image of the inner surface of fingers taking place between the upper phalanx (below the fingerprint) and the lower knuckle (the base of the finger), as shown in Fig.~\ref{fig:FTs_locations}. Its patterns are unique and reliable to be considered as a biometric characteristic \cite{Al-Nima2017Signal}. %\hl{The FTs can be found on the inner surface of five fingers: the four fingers or main fingers (little, ring, middle and index), and the thumb. \marginpar{This highlighted part is not necessary.}}  

%	\begin{figure}[!b]
%		\centering
%		\includegraphics[page=1,scale=.51,trim=0cm 0cm .6cm 3cm,clip]{Finger_biometrics.pdf}
%		\caption{Different physiological characteristics that can be found in each single finger}
%		\label{fig:Physiological_biometrics}
%	\end{figure}

%In the case of the FT literature review, few works have utilised the FT as a biometric type. The idea of using this important biometric only appeared approximately one decade ago. Because an FT recognition system is generally compromised within the following steps: finger segmentation and FT collection; feature extraction; matching classifier (multi-object fusion could be used in this stage) and recognition performance, related work will be reviewed for each step. 

The FT area involves the phalanx and knuckle patterns \cite{al2019segmenting}. The full FT region includes three phalanxes and three knuckles on the little finger, the ring finger, the middle and index fingers, and two phalanxes and two knuckles on the inner surface of the thumb \cite{Al-Nima2017Signal}. 

The three types of phalanxes for the four fingers are: (1) upper phalanx (below the fingerprint) called \textit{distal}; (2) middle phalanx named \textit{intermediate}; and (3) lower phalanx termed \textit{proximal}.

The three types of knuckles for the four fingers are: (1) \textit{upper knuckle} between the distal and intermediate phalanxes; (2) \textit{middle knuckle} between the intermediate and proximal phalanxes; and (3) \textit{lower knuckle} near the palm.

The two types of phalanxes for the thumb are: (1) upper phalanx (below the fingerprint) called \textit{distal} and (2) lower phalanx termed \textit{proximal}.

The two types of knuckles for the thumb are: (1) \textit{upper knuckle} between the distal and intermediate phalanxes and (2) \textit{lower knuckle} near the palm.

\begin{figure}[!h]
	\centering
	\includegraphics[page=1,scale=1,trim=4.3cm 14.5cm 4.3cm 3cm,clip]{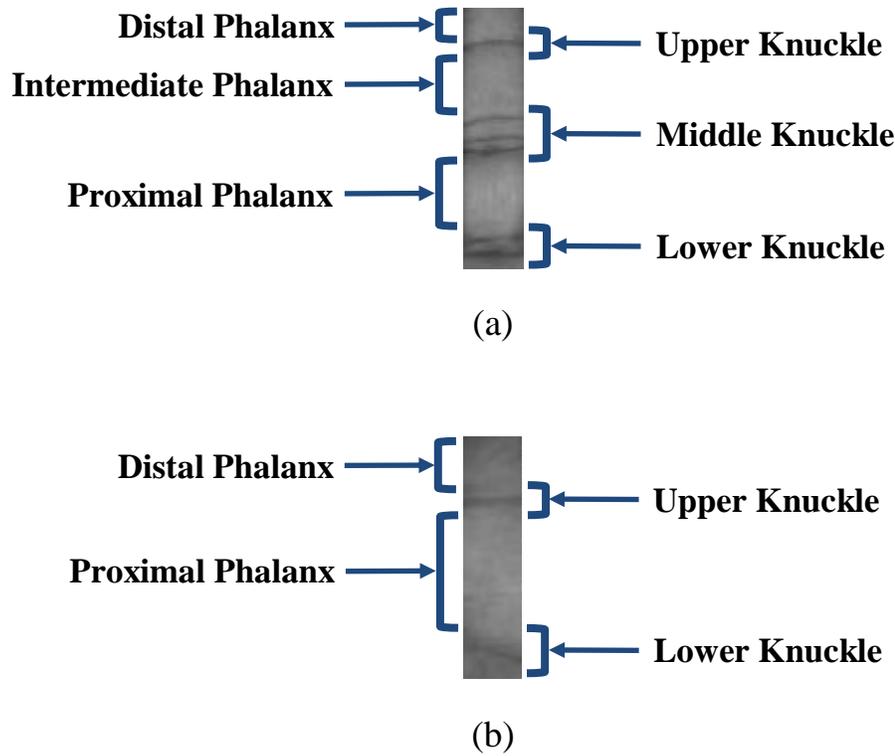}
	\caption{Example of the full FT parts in:
		(a) a middle finger (three phalanxes and three knuckles)
		(b) a thumb (two phalanxes and two knuckles)}	\label{fig:FT_midd_thumb}
\end{figure}

FT patterns are structured before the birth. The FTs can provide high recognition performance due to their different patterns in their various parts. The principal parts of the FT for a single finger are demonstrated in Fig. \ref{fig:FT_midd_thumb}.
%All the FT patterns (excluding the ridges) can be obtained from low cost devices and contactless hands. \\

\paragraph{Aim and structure of the paper} The aim of this paper is (1) to provide a comprehensive survey of the FT phenomenon in the case of biometric recognition; (2) to determine the main advantages and drawbacks of prior FT studies and the available FT databases; (3) to suggest insightful future directions to further improve the FT work.  

The rest of this paper is organised as follows: Section 2 describes the specifications of the finger characteristics
%	and provides comparisons with the FT
%; Section 3 highlights the main recognition system stages that were considered in the FT studies
; Section 3 illustrates the finger segmentation and FT collection studies; Section 4 states the employed FT feature extraction techniques; Section 5 reviews the multi-object fusion of FT work; 
%Section 7 explains the utilised and available FT databases; 
Section 6 surveys FT recognition performances; and finally Section 7 concludes the paper by presenting suggestions for future studies.

%%%%%%%%%%%%%%%%%%%%%%%%%%%%%%%%%%%%%%%%%%%%%%%%%%%%%%%%%%%%%%%%%%%%%%%%%%%%%%%%%%%%%%%%

\section{FT 
	%	Anatomy and Comparison with Other Physiological Finger 
	Characteristics}
As mentioned, the FT is located in the inner surface of a finger between the fingerprint and palm. It consists of knuckles and phalanxes. Moreover, it involves various types of patterns, these are vertical patterns (the visible lines); horizontal lines (the wrinkles) and ridges.

There are many facilities that encourage the use of the FT as a powerful biometric - they have rich information; they are unique for each person or even identical twins; easy to access; they can be acquired without contact; they are resistant to tiredness and emotional feelings; their main features are reliable and stable \cite{Bhaskar2014Hand}; they require an inexpensive scanner or camera to capture their images; they are located inside the fist, so, they are always protected \cite{Michael2010Robust}; precise recognition decision can be obtained by the cooperation of fingers and it has been noticed that the FTs will not change over life, even for people who play tennis, where such people need to use this part of their body to grasp the racket \cite{Ribaric2005ABiometric}. Also, one of the most interesting attributes in any finger characteristic is that its features are different not just between the individuals, in fact, among the fingers too.  

There are some issues that affect the FT in a biometric system, for instance, skin disease; injury or injuries; unclean inner finger surface and amputating a part or full finger. Nevertheless, the amputating issue was considered in \cite{Al-Nima2017Robust,Al-Nima2017finger}. That is, amputating one phalanx; two phalanxes; one finger and two fingers were considered in \cite{Al-Nima2017Robust}, and amputating one finger; two fingers and three fingers were studied in \cite{Al-Nima2017finger}.

\section{FT Recognition Systems}
First of all, biometric systems are generally recognized as identification systems or verification systems. In fact, a biometric system can be designed to work in one of three modes as will be detailed below:

$\bullet$ \textbf{Enrolment mode:} this is an initial step of any biometric system, where a template is created by storing features of the enrolled biometric characteristics. The enrolment mode is working as one-after-one policy. That is, each input has to be treated separately and independently starting from the capturing or scanning step and ending with distinctly saving its extracted features in the template. 

$\bullet$ \textbf{Verification mode:} in this mode a user claims his/her identity. Similar operations such as biometric acquiring, pre-processing and feature extraction are implemented. The resulting feature vector will then be matched with the same claimed identity vector. This module is known as one-to-one matching. Finally, the identity decision is to confirm or reject the claim. 
%	\begin{figure}[!t]
%		\begin{subfigure}[!t]{1\textwidth}
%			\centering
%			\includegraphics[page=1,height=2cm,width=7cm,trim=0cm 20.2cm 0cm 4cm,clip]{identification_verification_enrolment.pdf}
%			\caption{Enrolment mode, there are no comparisons in this mode as it just to store the features of the current enrolment input inside the template}
%			\label{subfig:Enrolment mode}
%		\end{subfigure}
%		\begin{subfigure}[!t]{1\textwidth}
%			\centering
%			\includegraphics[page=1,height=2cm,width=7cm, trim=0cm 12cm 0cm 9.5cm,clip]{identification_verification_enrolment.pdf}
%			\caption{Verification mode, in this mode a comparison is achieved by matching the features of the current input with the features of a claimed stored vector inside the template}
%			\label{subfig:Verification mode}
%		\end{subfigure}
%		\begin{subfigure}[!t]{1\textwidth}
%			\centering
%			\includegraphics[page=1,height=2cm,width=7cm, trim=0cm 4cm 0cm 17.5cm,clip]{identification_verification_enrolment.pdf}
%			\caption{Identification mode, in this mode a comparison is achieved by matching the feature vector of the current input with all of the stored feature vectors inside the template}
%			\label{subfig:Identification mode}
%		\end{subfigure}
%		\caption{The main operations of the enrolment, verification and identification modes. The parameter $N$ represents the total number of the processed biometric enrolment vectors and the parameter $Q$ represents a specific stored vector in the template}
%		\label{fig:enrolment_verification_identification}
%	\end{figure}

$\bullet$ \textbf{Identification mode:} like the prior operations of the verification and enrolment modes, the identification will extract the feature of the input biometric trait. However, a one-to-many matching will be implemented between the extracted feature vector and all of the stored vectors in the template. In this mode, the user cannot provide his identity. A decision is established by assigning the identity of the user or rejecting his/her membership of the biometric system. 

FT biometric systems were established in different approaches. Many papers presented the FT as a supported biometric. On the other hand, recent publications provided exhaustive FT studies such as \cite{Al-Nima2017Robust,Al-Nima2017finger}. In this survey, related FT work will be reviewed. In general, three stages were considered in the publications of FT recognition systems \cite{Al-Nima2017Robust,Al-Nima2017efficient}: %,Al-Nima2017Signal}: 
finger segmentation and FT collection, feature extraction and multi-object fusion.

The first stage to be concentrated on is the FT area. Although many studies have utilised the FT region, the full FT area was not always exploited. Furthermore, all the five fingers were not always employed. 

The second stage is the feature extraction, which can be considered as one of the most significant processes in any biometric system. This survey focused on this essential part to present the employed FT feature extraction methods. Several works exploited feature extraction techniques, but others suggested new approaches. 

The third stage is for the multi-object fusion. Some papers used the FT characteristic as a part of a combination biometrics system, where other biometric characteristics were utilised. On the other hand, recent publications concentrated on performing the fusion between just the FTs of finger objects. 

It can be observed that not all the stages were always utilised in the FT characteristic studies. The techniques that are used in each stage will be surveyed.

%%%%%%%%%%%%%%%%%%%%%%%%%%%%%%%%%%%%%%%%%%%%%%%%%%%%%%%%%%%%%%%%%%%%%%%%%%%%%%%%%%%%%%%%%
\section{Segmenting the FT Regions}
\label{Sec:Seg}
Segmenting the FT Regions is the first stage to be considered. Although the inner finger surfaces were exploited by many biometric papers, the full region of the FT was not always considered. The majority of the early work employed part of the area as in \cite{Ribaric2005Anonline,Ribaric2005ABiometric,Ferrer2007Low,ying2007identity,Pavesic2009Finger-based}. On the other hand, the recent studies have focused on exploiting all the FT parts of fingers such as \cite{Al-Nima2015Human,Al-Nima2017Robust,Al-Nima2017efficient}. Similarly, the full number of fingers was not always used. That is, some publications utilised four fingers and ignored the thumbs \cite{Ribaric2005Anonline,Ferrer2007Low,Pavesic2009Finger-based,michael2010innovative,Kanhangad2011AUnified,Al-Nima2015Human}. Other publications used all the five fingers of a hand image \cite{Ribaric2005ABiometric,ying2007identity,Goh2010Bi-modal,Michael2010Robust,Al-Nima2017Robust,Al-Nima2017efficient}. Nevertheless, several works used small parts of the FT region for only one or two fingers \cite{kumar2011contactless,Kumar2012Human,zhang2012hand}.
%		,stein2013video,sankaran2015Onsmartphone
%,malhotra2017fingerphoto}.

To breakdown the literature for segmenting FT regions, they can be partitioned into three categories: the studies that utilised few parts of the FT (small FT region). Secondly, the studies that exploited the majority of FT parts (big FT region). Thirdly, the studies that used all the FT parts (full FT region). Detailed information are illustrated as follows:
\paragraph{\textbf{Small FT region}} The distal phalanx with parts of upper and lower knuckles of the four fingers were neglected in \cite{Pavesic2009Finger-based}, where a study to fuse the fingerprints with the FTs of four hand fingers was established. Only index fingers were used in \cite{kumar2011contactless}, whilst, index and middle fingers were employed in \cite{Kumar2012Human}. These work used databases that have very small FT areas where many parts of the FT were ignored. These are the middle knuckle; proximal phalanx and lower knuckle.

\paragraph{\textbf{Big FT region}} The idea of employing the FTs started in \cite{Ribaric2005Anonline,Ribaric2005ABiometric}. A fixed ratio size of the FT was considered in each finger, and the lower knuckles were not fully involved. Only four fingers were used in \cite{Ribaric2005Anonline}, but all the five fingers were employed in \cite{Ribaric2005ABiometric}. A low cost fusion system between the palm, hand geometry and FTs was proposed in \cite{Ferrer2007Low}. The ROIs here were the full finger images except parts of the lower knuckle patterns, which were ignored. The thumb was not considered too. 

A segmentation method called \textit{Delaunay triangulation} for assigning hand parts was adopted. The key idea of this method is to simulate the hand image in a group of triangles to avoid the circular shapes \cite{ying2007identity}. This work discarded the lower knuckles and partially includes the proximal phalanx, where the ROI of a FT was determined by specifying 80\% from a finger area as cited in the same study. 

A finger segmentation method based on the projectile approach was proposed, where a system was designed to track the five fingers from the hand video stream. In the projectile method a middle point at the finger base was specified, then, it moved in a ``zigzag" path hitting the finger borders until reaching the finger tip point. Hereafter, the FT area was determined by a rectangle covering the region between the finger base and the tip, where part of the lower knuckle was not included. It is worth highlighting that no normalized resizing was applied for the FT rectangle \cite{Michael2010Robust}. 

Similarly, the same segmentation method was presented in \cite{michael2010innovative,Goh2010Bi-modal}, however, these works were implemented for only four fingers. The main problem of this projectile approach is that it could not detect very small distortions, rotations and translations as reported in \cite{Goh2010Bi-modal}. 

A method to segment the four fingers was suggested based on the following steps: applying the Otsu threshold \cite{otsu1975threshold} for the binarization; employing opening morphological operations; implementing a simple contour; specifying tips and valley points for the four fingers by using the local minima and local maxima; determining four points in both sides of each finger to specify the finger orientations and then segmenting the FT regions for only four fingers \cite{Kanhangad2011AUnified}. The lower knuckles were discarded, where they could increase the recognition performance of the FT. 

Only the middle finger after specifying the tips and valleys points of a hand fingers was segmented. Simply, the middle finger image was segmented by cropping the area between the tip and two valley points around this finger. Then, the full middle finger image was treated as the FT region. Part of the lower knuckle was included. Low resolution FT images of $30 \times 90$ pixels were collected \cite{zhang2012hand}. One can argue that it is not worth to include the fingerprint with FT patterns. 

A contactless multiple finger segments study was presented in the case of verification. Index finger images acquired from different nationalities (Arabian, African, Sri Lankan, Indians, Malays, Europeans and Chinese) with multiple rotations (between $+$45 to $-$45 degree) and scaling (between $12-20$ cm) were employed. All the index finger area was considered except the lower knuckle, where it was partially included. Finger images were segmented into three regions from the three knuckles. Then, combinations between their features; their segments and their segments and features were considered \cite{MAC2018contactless} \cite{Jahan2018Contactless}.
\begin{table}[!b]
	\centering
	\caption{Comparisons between the descriptions of different segmented FT region methods}
	\label{Table:FT_region}
	\scalebox{0.8}{\begin{tabular}{|C{4cm}|C{2cm}|C{4cm}|C{4.5cm}|}
			\hline
			\textbf{Reference} & \textbf{Number of employed fingers} & \textbf{Partially included FT parts} & \textbf{Not included FT parts} \\ \hline
			Ribaric and Fratric \cite{Ribaric2005Anonline}	& 4 & Lower knuckle & --- \\ \hline
			Ribaric and Fratric \cite{Ribaric2005ABiometric}	& 5 & Lower knuckle & --- \\ \hline
			Ferrer \textit{et al.} \cite{Ferrer2007Low} & 4 & Lower knuckle & --- \\ \hline
			Ying \textit{et al.} \cite{ying2007identity} & 5 & Proximal phalanx & Lower knuckle \\ \hline
			Pavesic \textit{et al.} \cite{Pavesic2009Finger-based} & 4 & Upper knuckle and Lower knuckle & Distal phalanx \\ \hline
			Michael \textit{et al.} \cite{Michael2010Robust}	& 5 & Lower knuckle & --- \\ \hline
			Michael \textit{et al.} \cite{michael2010innovative} & 4 & Lower knuckle & --- \\ \hline
			Goh {et al.} \cite{Goh2010Bi-modal} & 4 & Lower knuckle & --- \\ \hline
			Kanhangad \textit{et al.} \cite{Kanhangad2011AUnified} & 4 & --- & Lower knuckle \\ \hline
			Kumar and Zhou \cite{kumar2011contactless} & 1 & Intermediate Phalanx & Middle knuckle, proximal phalanx, and lower knuckle \\ \hline
			A. Kumar and Y. Zhou \cite{Kumar2012Human} & 2 & Intermediate Phalanx & Middle knuckle, proximal phalanx, and lower knuckle \\ \hline
			Zhang \textit{et al.} \cite{zhang2012hand} & 1 & Lower knuckle & --- \\ \hline 
			%					Stein \textit{et al.} \cite{stein2013video} & 2 (one from each hand) & Intermediate Phalanx & Middle knuckle, proximal phalanx and lower knuckle \\ \hline
			%					Sankaran \textit{et al.} \cite{sankaran2015Onsmartphone} & 2 & Middle knuckle or proximal phalanx & Proximal phalanx and lower knuckle, or lower knuckle only \\ \hline
			Al-Nima \textit{et al.} \cite{Al-Nima2015Human} & 4 & --- & --- \\ \hline
			%					Malhotra \textit{et al.} \cite{malhotra2017fingerphoto} & 2 & Middle knuckle or proximal phalanx & Proximal phalanx and lower knuckle, or lower knuckle only \\ \hline
			Al-Nima \textit{et al.} \cite{Al-Nima2017Robust} & 5 & --- & --- \\ \hline
			Al-Nima \textit{et al.} \cite{Al-Nima2017efficient} & 5 & --- & --- \\ \hline
			%					Debayan \textit{et al.} \cite{Debayan2018matching} & 4 (two from each hand) & Upper knuckle & Intermediate phalanx, middle knuckle, proximal phalanx and lower knuckle \\ \hline
			MAC \textit{et al.} \cite{MAC2018contactless} & 1 index (Not clear from 1 or 2 hands)  & Lower knuckle & --- \\ \hline
			MAC \textit{et al.} \cite{Jahan2018Contactless} & 1 index (Not clear from 1 or 2 hands)  & Lower knuckle & --- \\ \hline
			%					Wasnik \textit{et al.} \cite{Wasnik2018Improved} & 1 (from left hand) & Intermediate phalanx & Middle knuckle; proximal phalanx and lower knuckle \\ \hline
			%					Wasnik \textit{et al.}\cite{wasnik2018baseline} & 1 (from left hand) & Intermediate phalanx & Middle knuckle; proximal phalanx and lower knuckle \\ \hline
			%					\multirow{4}{*}{Weissenfeld \textit{et al.} \cite{Weissenfeld2018contactless}} & 4 & Intermediate phalanx & Middle knuckle; proximal phalanx and lower knuckle \\ \cline{2-4}
			%					& 2 (1 index from each hand) & Intermediate phalanx & Middle knuckle; proximal phalanx and lower knuckle \\ 
			%					\hline
			
			%					Chopra \textit{et al.} \cite{Chopra2018Unconstrained} & 2 (separetely or together) & Adaptively partially included parts & Adaptively excluded parts \\ 
			%					\hline
	\end{tabular}}
\end{table}

\paragraph{\textbf{Full FT region}} A method to segment all the FT parts from the four fingers was suggested. This publication confirmed that including the lower knuckle patterns within the FT region would increase the performance of the biometric recognition. That is, the Equal Error Rates (EERs) after adding the third or lower knuckles recorded better results than the EERs without these important features such as the EER percentage has been reduced from 5.42\% to 4.07\% by using a feature extraction method termed the Image Feature Enhancement (IFE) based exponential histogram and it has been reduced from 12.66\% to 7.01\% by using the IFE based bell-shaped histogram \cite{Al-Nima2015Human}. Effectively, the main FT regions for the four fingers were assigned in this study, but the FT of the thumb was not included.
%In \cite{Al-Nima2015Human}, it showed that the EER was reduced from 5.42\% to 4.07\% by adding the third or lower knuckles and using a feature extraction method - the Image Feature Enhancement (IFE) based on exponential histogram. 
%The EER was reduced from 12.66\% to 7.01\% by using the IFE based bell shaped histogram. 
%Effectively, the main FT regions for the four fingers were assigned in this study, but the FT of the thumb was not included.
%The Hong Kong Polytechnic University Contact-free 3D/2D (PolyU3D2D) Hand Images Database (Version 1.0) \cite{Databasever1PolyU3D2D} was used. 

A robust approach for the finger segmentation was proposed in \cite{Al-Nima2017Robust}. This segmentation method considered each finger as an object. It maintained the hand image before carrying out the segmentation process. To explain, multiple image processing operations were adopted. The images of the five fingers were collected from a large number of contact free hand image databases. This was followed by defining a ROI of each FT. An adaptive inner rectangle was utilised to segment the ROIs of the five fingers. The suggested finger segmentation was appropriate for contactless hand images, where it could efficiently manage the translations and scaling of hand images. The FT parts were fully employed in this study. 

An adaptive and robust finger segmentation method was proposed to solve the problem of a hand alignment variation. As such, it could be adapted to different hand alignments such as rotations and translations. A scanning line was suggested to detect the hand position and determine the main specifications of the fingers. Furthermore, an adaptive threshold and adaptive rotation step were exploited. The proposed segmentation approach could carry out the various degrees of translations, scalings and orientations. All the FT parts were used \cite{Al-Nima2017efficient}. 

A comparison has been established for the segmented FT regions methods as illustrated in Table \ref{Table:FT_region}. To summarize, the majority of the work in the FT area only explored part of the finger texture areas for recognition. This issue has been addressed recently, but only in a few studies. It can be argued that the achieved recognition performance can be enhanced if more patterns of the FTs are included \cite{Al-Nima2015Human,Al-Nima2017Robust,Al-Nima2017efficient,Al-Nima2017finger}.%,Al-Nima2017Signal}.

\section{FT Feature Extractions}
\label{sub:feature_extraction}
Feature extraction is one of the most important parts in any biometric system. This aspect is considered for the FT in this survey. Three types of FT patterns can be found: vertical lines, horizontal lines and ridges. The vertical and horizontal lines can be considered as the main patterns of FT. They can be collected by using inexpensive and low resolution capturing equipments. Whereas, the ridge pattern needs high resolution acquiring devices. In the case of feature extractions, known methods were applied such as the Haar wavelet% \cite{Haar_Wavelet}
, Principal Component Analysis (PCA), Ridgelet transform% \cite{Ridgelet_Transform}
and Competitive Coding (CompCode). Furthermore, more efficient feature extractions which have specifically been designed for the FT patterns were employed such as the Localized Radon transform (LRT), Scattering Networks (ScatNet), Enhanced Local Line Binary Pattern (ELLBP) and Multi-scale Sobel Angles Local Binary Pattern (MSALBP). 

FT feature extraction literature can be divided into three groups: some considered the general FT features \cite{Ribaric2005Anonline,Ribaric2005ABiometric,Ferrer2007Low,Pavesic2009Finger-based,Zhang2010hand,Kanhangad2011AUnified,zhang2012hand,Al-Nima2015Human,Al-Nima2016ANovel,omar2018deep}, others concentrated on vertical and horizontal lines patterns \cite{ying2007identity,nanni2009multi,Michael2010Robust,michael2010innovative,Goh2010Bi-modal,Bhaskar2014Hand,Al-Nima2017Robust,Al-Nima2017efficient,Al-Nima2017finger,al2018personal,al2018finger} and those focused on ridge patterns \cite{kumar2011contactless,Kumar2012Human,MAC2018contactless,Jahan2018Contactless}. These groups of work are explained as follows:
\paragraph{\textbf{General FT features}} A feature extraction technique based on the eigenvalues was introduced by \cite{Ribaric2005Anonline,Ribaric2005ABiometric}. This method was applied as a feature extraction to the FTs and produced eigenfingers. In \cite{Ribaric2005ABiometric}, this method was also applied for the palmprints and generated eigenpalms. Only the most important features of produced eigenvector according to a determined eigenvalue are collected% (supported matlab code can be found in \cite{Eigenvalues_Eigenvectors})
. The challenge here is choosing the best eigenvalue - low values tend to ignore some features and high values would collect noises as explained by the authors. 

A feature extraction based on encoding schemes of various 2D Gabor phase %\cite{gabor_filter} 
to analyse the texture of the FTs and palmprint was used. Then, each FT was binarized to a number between 1 and 3000 for each pixel. The resulting images were used as FT features and they contained only global patterns (not textures). Subsequently, four featured images for the four fingers were concatenated after the binarization process. The extracted features of FT images were represented by binary flat areas \cite{Ferrer2007Low}. Obviously, these features are weak compared to the real FT features such as horizontal lines, vertical lines and ridges. 

A combination system between fingerprints and FTs of the four fingers was generated. Three feature extraction methods were evaluated in this work: PCA, the Most Discriminant Features (MDF) and the Regularized-Direct Linear Discriminant Analysis (RDLDA). The best results were reported for the RDLDA method \cite{Pavesic2009Finger-based}. Extracting the discriminant features was the target of the feature extraction methods, but choosing the appropriate parameters for each method to avoid collecting image noises appears to be a big problem. 

A combination between the features of only the middle finger and the palmprint in one Single Sample Biometrics Recognition (SSBR) system was applied. The segmented middle finger area was treated for the FT region. A Locality Preserving Projections (LPP) transform was implemented as a feature extraction to both middle finger and palmprint images. Normalization computations were employed on the results. Subsequently, a PCA was used to preserve the fusion feature and reduce the information size. As illustrated by the authors, the LPP feature extraction could obtain the main discriminant features (or essential structures of the FT patterns) \cite{Zhang2010hand}, but wastes other non-discriminant features. 

A CompCode method as a feature extraction was utilised by \cite{Kanhangad2011AUnified}. The authors also utilised a Hamming Distance (HD) as a matching metric between the templates and the testing vectors. The CompCode method was approached in \cite{Kong2004Competitive} by applying competitive codes to multiply 2-D Gabor filters in order to extract the rotation features. Only 6 values were exploited to represent the extracted features and this is not sufficient to describe the variances between different patterns. 

For the work in \cite{zhang2012hand}, a fusion between the palmprint and the middle finger was performed. The segmented middle finger region was exploited for the FT area as in \cite{Zhang2010hand}. An LPP was employed for each two dimensional wavelet feature of both biometrics. The sub-band wavelet coefficients of approximation, horizontal details and vertical details were separately collected for each biometric. An average filter was applied just for the horizontal and vertical details of the palmprint coefficients. Then, the LPP methods were applied to each sub-band wavelet. Discriminant features of approximation, horizontal details and vertical details were extracted in this study and other features were excluded. 

%	A fingerphoto verification algorithm that used a smartphone camera was designed by \cite{sankaran2015Onsmartphone}, where the fingerprint was used with the FT. The fingerphoto image was firstly enhanced after converting to the grayscale as follows: employing the median filter, applying Histogram equalization and performing the sharpening operation. Subsequently, a novel ScatNet method was described for the feature extraction. It basically consists of a filter bank of wavelets. It can generate unchanging pattern representation. General minutiae features were obtained. Therefore, all other FT features were wasted as micro-texture features. 

A feature extraction method named Image Feature Enhancement (IFE) was employed in \cite{Al-Nima2015Human}. It includes image processing operations. These operations are the Contrast Limited Adaptive
Histogram Equalization (CLAHE), for adjusting the brightness of the FT, and a contrast feature fusion. The contrast feature fusion involves extracting the lower information of the CLAHE image; subtracting the resulted values from the CLAHE image; extracting the upper information from the CLAHE image and adding them to the resulting subtracted image. Three types of CLAHE histogram distributions were investigated - bell-shaped, exponential and flat histograms. Experimental results highlighted that the exponential distributions histogram achieved the best performance \cite{Al-Nima2015Human}. Discriminant FT features could be extracted here, whilst other non-discriminant features were ignored. 

Three feature extraction methods were assessed in \cite{Al-Nima2016ANovel}: a statistical calculation named Coefficient of Variance (CV); Gabor filter with the CV and Local Binary Pattern (LBP) with the CV. The aim of this work is to establish Receiver Operating Characteristic (ROC) graphs for the Probabilistic Neural Networks (PNNs) by proposing a novel approach. The best result was obtained by using the LBP with the CV \cite{Al-Nima2016ANovel}. This is because that the LBP with the CV could analyse texture FT features, whereas the Gabor filter with the CV and only the CV could extract general FT features. Noise problems affect the first feature extraction method. 

%	The same ScatNet feature extraction as in \cite{sankaran2015Onsmartphone} was used by \cite{malhotra2017fingerphoto}, where the latter can be considered as an extended work. However, another enhancement method based on the LBP was presented. As mentioned, the ScatNet method can extract the general minutiae information. Whilst, other features were avoided such as the micro-textures. 

%	A baseline comparison work for a fingerphoto verification application in a mobile phone was provided. Three feature extraction methods were investigated. These are the LBP; Histogram of Oriented Gradients (HOG) and Binarized Statistical Image Features (BSIF). The authors compared the three feature extraction methods with the Commercial Off-The-Shelf (COTS) method. The COTS obtained higher performance than other feature extraction. Therefore, the authors advised to use advanced pre-processing technique and overcome the commercial applications \cite{wasnik2018baseline}. 

%	In \cite{Chopra2018Unconstrained}, two feature extraction methods for unconstrained fingerphoto images were used. These are the CompCode \cite{Kong2004Competitive} with a HD and the ResNet50 \cite{He2016Deep} with the cosine, the HD and cosine were used to measure the similarity between the compared images. Both of the two feature extraction methods attained low performance as the key idea of this work was attaining high segmentation accuracy.

The deep learning was exploited with the FT for people authentication. A novel Deep Finger Texture Learning (DFTL) network is established in this work and this can be considered as the first approach that employed deep learning in a FT study. According to the deep learning concept, the DFTL handles extracting the FT features during the training phase by using the backpropagation method \cite{omar2018deep}. The main problem here is that the feature extraction in the DFTL is based on the trained samples only and it could not be appropriate for some testing inputs. 

\paragraph{\textbf{Vertical and horizontal lines patterns}} A holistic feature extraction method was proposed by \cite{ying2007identity}. It consists of the following operations: denoting landmarks points of the geometrical information of a hand image; employing the image warping filter to remap the geometrical information; applying the binarization on textures and using the HD to measure the similarities. An holistic method was applied to all hand parts (palm and fingers). As explained in this publication, the extracted features are mainly the horizontal and vertical lines because these features preserve their permanent locations after applying the warping filter, on the other hand, this feature extraction is not robust to recognize dislocating or scaling patterns. 

Two feature extraction methods were used in \cite{nanni2009multi}. These are the Radon transform% \cite{Radon_Transform}
and the Haar wavelet. The outcome of each feature extraction method was transformed by using the non-linear Fisher transform. Consequently, a score fusion was applied for the resulting values. This work focused on only vertical line details and ignored other features such as horizontal line patterns. The ridgelet transform was selected by \cite{Michael2010Robust}, \cite{michael2010innovative} and \cite{Goh2010Bi-modal} to be applied as a feature extraction. The ridgelet transform is fundamentally constructed for images with lines. This makes it suitable for analysing the main FT patterns of horizontal lines (or wrinkles) and vertical lines (or knuckles). The essential advantage of the ridgelet method is its ability to collect the line patterns. However, the important features of micro-textures \cite{al2017multi} are ignored and this is the big drawback of the ridgelet. 

Two FT feature extractions - the Scale Invariant Feature Transform (SIFT) %\cite{SIFT_FeatureExtreaction} 
and Ridgelet transform - were evaluated in \cite{Bhaskar2014Hand}. This work recorded that the SIFT obtained better results than the Ridgelet transform. According to this paper, only the line patterns were extracted. All other features were ignored. In \cite{Al-Nima2017Robust}, a feature extraction enhancement called the ELLBP was proposed. It is an enhanced version of the Local Line Binary Pattern (LLBP) \cite{Petpon2009Face%,Efficient_LLBP
}. It is based on fusing the main FT patterns of horizontal and vertical textures by employing the weighted summation rule, which was found to be beneficial to describe the main FT patterns. 

Choosing the fusion parameters (or horizontal and vertical weights) is not a straightforward task. The authors in \cite{Al-Nima2017Robust} partitioned the training samples into training and validation subsets to determine the values of the fusion parameters. LLBP feature extraction was exploited to analyse the horizontal and vertical patterns \cite{Al-Nima2017efficient}. The main problem in the LLBP approach can be found in its amplitude fusion, where the amplitude computations are not appropriate to provide directional information. They can be influenced by noise, brightness and range value according to \cite{young1998fundamentals}, therefore, it cannot give effective description of image textures. 

A novel FT feature extraction method termed the MSALBP was illustrated. Briefly, the MSALBP approach consists of the following operations: obtaining the Sobel horizontal and vertical edges of the FT; combining them according to their directional angles; fusing the resulted image with the Multi-Scale Local Binary Pattern (MSLBP); partitioning the outcome values into non-overlapping windows and performing the statistical calculations to produce a texture vector \cite{Al-Nima2017finger}. The main drawback here is that multiple operations were combined in this feature extraction which resulted in increasing the complexity of this method . 

A new feature extraction approach named the Surrounded Patterns Code (SPC) was proposed. This method was utilised to collect the	surrounding patterns near the vertical and horizontal lines of FTs. This method analyses the surrounding patterns of vertical and horizontal lines separately. Then, it combines the obtained surrounded pattern values \cite{al2018personal}. Although, the SPC analyses well the surrounding features, it ignores the main FT patterns of vertical and horizontal lines.

For the study of \cite{al2018finger}, the ELLBP method was utilised to extract the FT features of four fingers. In this paper, the ELLBP was applied to the CASIAMS (Spectral 460) and CASIAMS (Spectral White) databases. Again, the ELLBP has the capability of analysing the main FT patterns even for the different FTs that are acquired under different spectra. The authors examined various fusion methods between the FT features of two CASIAMS spectra.

\begin{table}[!b]
	\centering
	\caption{Summary of employed feature extraction methods by the related FT studies}
	\label{Table:feature_extraction}
	\scalebox{0.7}{\begin{tabular}{|C{3.5cm}|C{3.5cm}|C{4cm}|C{5.5cm}|}
			\hline
			\textbf{Reference} &  \textbf{Feature extraction method} & \textbf{Extracted features} & \textbf{Feature extraction drawback} \\ \hline
			Ribaric and Fratric \cite{Ribaric2005Anonline}		&  Eigenfingers & Most important
			features of eigenvectors & Choosing best eigenvalue\\ \hline
			Ribaric and Fratric \cite{Ribaric2005ABiometric}		&  Eigenfingers & Most important
			features of eigenvectors & Choosing best eigenvalue\\ \hline
			Ferrer \textit{et al.} \cite{Ferrer2007Low} 		& 2D Gabor phase encoding scheme & Binary flat areas in FT images & Features of binary flat areas are weak \\ \hline
			Ying \textit{et al.} \cite{ying2007identity} 		&  Holistic method & Mainly horizontal and vertical lines & Not robust to dislocating or scaling patterns\\ \hline
			Nanni and Lumini \cite{nanni2009multi} 		&  Radon transform and Haar wavelet & Vertical lines details & Ignoring other features as horizontal lines\\ \hline
			Pavesic \textit{et al.} \cite{Pavesic2009Finger-based} 		&  PCA, MDF and RDLDA & Discriminant features & Choosing best tuned parameters\\ \hline
			Michael \textit{et al.} \cite{Michael2010Robust}		& Ridgelet transform & Line patterns & Ignoring micro-texture features\\ \hline
			Michael \textit{et al.} \cite{michael2010innovative} 		& Ridgelet transform & Line patterns & Ignoring micro-texture features\\ \hline
			Goh \textit{et al.} \cite{Goh2010Bi-modal} 		& Ridgelet transform & Line patterns & Ignoring micro-texture features\\ \hline
			Zhang \textit{et al.} \cite{Zhang2010hand} & LPP transform & General discriminant features & Ignoring non-discriminant features\\ \hline
			Kanhangad \textit{et al.} \cite{Kanhangad2011AUnified} 		& CompCode & Lines orientation information & Using only 6 values to represent the features \\ \hline
			Kumar and Zhou \cite{kumar2011contactless} 	& CompCode and LRT & Lines and curves orientation information & Using few values to represent the features \\ \hline
			A. Kumar and Y. Zhou \cite{Kumar2012Human} 		& CompCode and LRT & Lines and curves orientation information & Using few values to represent the features \\ \hline
			Zhang \textit{et al.} \cite{zhang2012hand} 		& LPP based on 2D wavelet transform & Discriminant features of approximation, horizontal details and vertical details & Ignoring non-discriminant features\\ \hline
			%			Stein \textit{et al.} \cite{stein2013video} 		& Median-filter + adaptive threshold & Minutiae information (binary) & Using binary representation is weak\\ \hline
			Bhaskar and Veluchamy \cite{Bhaskar2014Hand} 		& Ridgelet transform and SIFT & Line patterns & Ignoring micro-texture features\\ \hline

			%			Sankaran \textit{et al.} \cite{sankaran2015Onsmartphone} 		& ScatNet & General minutiae features & Ignoring micro-texture features\\ \hline
			Al-Nima \textit{et al.} \cite{Al-Nima2015Human} 		& IFE & Discriminant features & Ignoring non-discriminant features\\ \hline
			Al-Nima \textit{et al.} \cite{Al-Nima2016ANovel} 		& LBP+CV, Gabor filter+CV and only CV & Texture features by LBP+CV and general features by Gabor filter+CV and only CV & Noise problems for LBP+CV and ignoring micro-texture features for Gabor filter+CV and only CV \\ \hline
			%			Malhotra \textit{et al.} \cite{malhotra2017fingerphoto} 		& ScatNet & General minutiae features & Ignoring micro-texture features\\ \hline
			Al-Nima \textit{et al.} \cite{Al-Nima2017Robust} 		& ELLBP & Horizontal and vertical lines & Choosing fusion parameters of horizontal and vertical weights \\ \hline
			Al-Nima \textit{et al.} \cite{Al-Nima2017efficient} 	& LLBP & Horizontal and vertical lines & Resulting high values by amplitude fusion \\ \hline
			Al-Nima \textit{et al.} \cite{Al-Nima2017finger} 		& MSALBP & Horizontal and vertical lines & Using multiple combination operations\\ \hline
			MAC \textit{et al.} \cite{MAC2018contactless} & HOS of ridge orientation & Ridge orientation & The line patterns of knuckles were ignored \\ \hline
			MAC \textit{et al.} \cite{Jahan2018Contactless} & HOS of ridge orientation & Ridge orientation & The line patterns of knuckles were ignored \\ \hline
			%			Wasnik \textit{et al.} \cite{Wasnik2018Improved} & multi-scale second order Gaussian derivatives & Ridge local structures & Selecting ($sigma$) parameter \\ \hline
			
			%			Wasnik \textit{et al.} \cite{wasnik2018baseline} & LBP; HOG and BSIF & Different features for comparisons & COST overcome applied feature extractions \\ \hline
			
			%			Chopra \textit{et al.} \cite{Chopra2018Unconstrained} & CompCode and ResNet50 & General ridge features & Reported low performances \\ \hline
			
			Omar \textit{et al.} \cite{omar2018deep} & DFTL & Exclusive features using the deep learning & Based on training samples only \\ \hline
			Al-Nima \textit{et al.} \cite{al2018personal} & SPC & Surrounded patterns around the vertical and horizontal lines & Ignoring the main FT patterns of vertical and horizontal lines \\ \hline
			Al-Kaltakchi \textit{et al.} \cite{al2018finger} & ELLBP & Horizontal and vertical lines & Choosing fusion parameters of horizontal and vertical weights \\ \hline
	\end{tabular}}
\end{table}
\paragraph{\textbf{Ridge patterns}} In \cite{kumar2011contactless,Kumar2012Human} two FT feature extraction methods were observed: the Gabor-Filter-Based Orientation Encoding, or CompCode, and the LRT. It was found that the LRT attained better performance than the CompCode. Both feature extraction methods concentrate on collecting the orientation information of the lines and curves patterns. Again, few values were utilised to represent the features and this is not enough to describe the variances between various patterns. As mentioned, very small regions of FTs were used in these studies. 
%	Fingerphotos recognition and anti-spoofing methods were produced by \cite{stein2013video}. The fingerphoto images were acquired by using mobile-phone cameras. As mentioned in Section \ref{Sec:Seg}, part of FT regions from only right and left index fingers were utilised for each subject, where two areas were determined ``core area" and ``outer area". The recognition in this work was basically designed for the ``core area". So, for this area a simple feature extraction of Median-filter, kernel size $3 \times 3$, and adaptive threshold was used for binarization. Subsequently, the minutiae information were obtained in binary. Although the minutiae details were appeared, using binary values is weak to represent the features. 

A contactless multiple finger segments work was produced. Index finger images were acquired under different rotations and scaling. Higher order spectral (HOS) representations of ridge orientation feature extractions were employed \cite{MAC2018contactless,Jahan2018Contactless}. Nonetheless, the line patterns of knuckles were ignored. 

%	An improved fingerphoto verification system was designed as a new feature extraction method from the eigenvalues of convolved finger images utilising multi-scale second order Gaussian derivatives. This method was found to be useful for ridge patterns especially in recognizing fingerphotos by using a smartphone application. The problem of this method appears in selecting the Gaussian slope parameter ($sigma$). In the paper, the authors examined 10 values of $sigma$, then the maximum pixel values of the resulted image were considered \cite{Wasnik2018Improved}.

A summary of employed feature extraction methods by the related FT studies are given in Table \ref{Table:feature_extraction}. From this table, it can be observed that different types of feature extractions were employed. Many papers utilised established feature extraction methods and others proposed new approaches. In general, the appropriate FT feature extraction was found to be the one which can efficiently analyse the FT patterns \cite{kumar2011contactless,Kumar2012Human,
	%		sankaran2015Onsmartphone,
	Al-Nima2017Robust,Al-Nima2017efficient,Al-Nima2017finger}.%,Al-Nima2017Signal}. 

\section{Multi-Object Fusion}
For the multi-object biometric prototype, many publications have documented the FT as a part of multi-modal biometric recognition. A multi-modal biometric system is defined as a system that combines multiple characteristics in one biometric system \cite{ISO2012Information}. Later studies focused on designing multi-object biometric systems based on fusing the FTs of multiple fingers together to enhance the performance of a single-modal system. A single-modal biometric system is denoted as a system that employs only one biometric characteristic. Commonly, there are four levels of fusions: sensor level fusion; feature level fusion; score level fusion and decision level fusion \cite{almahafzah2012multibiometric}. A determined rule can be applied at each level such as summation rule, multiplication rule and weighted summation rule. 

To simplify the representation of the multi-object fusion literature, they are divided into three sets. Firstly, the work that combined two characteristic objects \cite{nanni2009multi,Michael2010Robust,michael2010innovative,Goh2010Bi-modal,Zhang2010hand,zhang2012hand,Kumar2012Human,MAC2018contactless,Jahan2018Contactless}.
%		,Debayan2018matching,Chopra2018Unconstrained,Badrinarayanan2017SegNet}.
Secondly, the work that fused multi-characteristic objects (three or more) \cite{Ribaric2005Anonline,Ribaric2005ABiometric,Ferrer2007Low,ying2007identity,Pavesic2009Finger-based,Kanhangad2011AUnified}. Thirdly, the work that used multi-FT objects \cite{Al-Nima2015Human,Al-Nima2016ANovel,Al-Nima2017Robust,Al-Nima2017efficient,Al-Nima2017finger,al2018personal,al2018finger}. These sets are highlighted as follows:
\paragraph{\textbf{Two characteristic objects}} A fusion of multiple matcher scores by utilising the summation rule was suggested in \cite{nanni2009multi}, where two different feature extraction processes were applied to the same FT, as described in Section \ref{sub:feature_extraction}. Each feature extraction process ended by a matcher operation. A fusion between the matching scores was then performed. The proposed approach was firstly implemented for only middle fingers. After that, ring fingers were augmented to improve the recognition performance. 

The same fusion method was presented between the FTs and the palmprint. Again, a score fusion was exploited, but by applying the Support Vector Machine (SVM) technique with the Radial Basis Function (RBF) kernel. Before that, a matching was applied by using the HD for the palmprint and Euclidean distance for the FTs \cite{Michael2010Robust,michael2010innovative}. Five fingers were used in \cite{Michael2010Robust}, whilst, the thumb was excluded in \cite{michael2010innovative}. 

Similarly, a robust recognition system for fusing the FTs of the five fingers with the palmprint in the case of verification was described by \cite{Goh2010Bi-modal}. In this study, the FTs were used as one subject to be combined with the palmprint. A score fusion was utilised between their matchers by applying different fusion rules: AND, OR, summation and weighted summation. The weighted summation rule was highlighted as a best choice. A feature fusion between only the middle finger and the palmprint was implemented by \cite{Zhang2010hand}. In this publication, a feature level fusion based on the concatenation rule was performed for their features, which were obtained by the LPP transform. Consequently, PCA was applied for the resulting fusion before using a nearest neighbour classifier for the recognition. 

Two level fusions in one SSBR system were suggested. They were considered between only the middle finger and the palmprint. Firstly, feature level fusions were performed for the wavelet coefficients of each trait by exploiting a weighted concatenation rule. Secondly, a score fusion level was executed by using the summation rule after the distance metric measures of each feature level fusion \cite{zhang2012hand}. Texture and vein images of only two fingers (middle and/or index) were evaluated in \cite{Kumar2012Human}. 
%in two experiments, where various numbers of people were applied for the personal identification. 
The main problem of the employed database in this publication is that it utilised a small area of FT. It can be observed that the authors were obtaining two innovative rules of a score fusion termed non-linear and holistic, both of these rules were based on the vein features. 

A contactless multiple finger segment objects was investigated. Index finger images were used and segmented into three regions from landmarks of the three finger knuckles, so, the fingerprints were included too. Consequently, fusions between the features of the segmented regions; the data of the segmented regions and both of them were studied \cite{MAC2018contactless}. In \cite{Jahan2018Contactless}, a very similar result as in the previous study was presented. That is, only a combination between the data of the segmented index finger regions was exploited.

%	Two Android applications for fingerphoto ridges images were constructed. Fingerprints with small parts of FTs of only two index fingers and two thumb fingers from each hand were considered.	Three fusion methods were applied after separately evaluating each finger, fusion between two thumbs or fusion between two index fingers; fusion between thumb and index fingers of each hand; and fusion of all employed fingers (thumb and index of both hands). Summation rules of score level combinations were applied to all utilised fusions. High performances could be obtained by employing fusion methods \cite{Debayan2018matching}.

%	For the study of \cite{Chopra2018Unconstrained}, a fingerphoto segmentation method utilising the VGG SegNet \cite{Badrinarayanan2017SegNet} was proposed. It is a pre-trained deep network and it was fine-tuned for the fingerphoto segmentation issue. Index and/or middle fingers from any hand were employed. The fingerprint region was included with the segmented finger image. Part of FT area might be excluded according to the segmentation process. 
\paragraph{\textbf{Multi-characteristic objects}} The first investigation here was for combining FTs with the finger-geometry in the case of human identification and verification \cite{Ribaric2005Anonline}. In this work, five templates were established, four templates for the extracted features of the four fingers and one template for the measurements of the fingers-geometry. Euclidean distance calculations were used between the corresponding templates. Then, the fusion was applied at the score level based on the weighted summation rule, which was named the total similarity measure. 

A method of fusing FTs with palmprints was designed in the case of human identification. In this publication, a score fusion for the similarity measures of the five FTs and the palmprint was implemented. This fusion was based on the weighted summation rule \cite{Ribaric2005ABiometric}. 

An inexpensive multi-modal biometric identification system by performing a fusion between the hand geometry, FTs and palmprint was generated in \cite{Ferrer2007Low}. Here, different types of combinations were examined: decision fusion with the voting rule, score level fusion by the weighted summation and feature fusion based on the two-dimensional convolution. It was experimentally cited that the decision fusion achieved the most satisfactory results. 

Two fusion methods between FTs of the five fingers and the palmprint were evaluated by \cite{ying2007identity}. The first method was based on the feature level fusion termed holistic as all the regions of the FTs and palmprint were exploited in the same feature extraction method. The second combination method was suggested to use a weighted summation rule as a type of score fusion, where hand parts (palmprint and FT of each single finger) were processed separately then the score fusion was achieved after applying the HD matcher. Finally, the score fusion was experimentally found to attain better results than the holistic method. 

Eight regions from the four fingers were determined to be used for human verification and identification. The two main considered characteristics were the digitprint, which represents the FT of each finger, and the fingerprint. After performing the feature extraction for each segmented region, an Euclidean distance matching module was implemented between the resulting values and the corresponding data in a template. A score fusion was employed for all the matching modules by applying the weighted summation rule \cite{Pavesic2009Finger-based}. 

Combinations of various hand characteristics were performed in \cite{Kanhangad2011AUnified}. Basically, these characteristics were FTs; hand geometry and palmprint. Furthermore, 2D and 3D biometric features were studied for the hand geometry and palmprint. Combinations between the distance matchers of the different biometric features were performed at score levels by utilising the weighted summation rule. 

\paragraph{\textbf{Multi-FT objects}} Feature level fusion based on the concatenation rule between the FT features of multiple finger objects was employed \cite{Al-Nima2015Human,Al-Nima2016ANovel,Al-Nima2017Robust,Al-Nima2017efficient}. More experiments were included in \cite{Al-Nima2017Robust} to examine the verification performance with missing finger elements (or parts). For example, removing a distal phalanx; a distal and intermediate phalanxes; one finger and two fingers. A salvage approach has also been suggested, implemented and analysed to increase the verification performance rates in the case of such missing elements by using the embedded information in a PNN.

Another combination method was explained as a novel neural network named the Finger Contribution Fusion Neural Network (FCFNN). The FCFNN fuses the contribution scores of the finger objects. This approach was inspired from the different contribution of each finger, where the contribution score of any finger in terms of individual verification is not the same as the contribution score of the other fingers. This neural network has an advantage in its flexible architecture. That is if any finger is accidentally amputated, is easily to be ignored by removing its connections from the network. So, removing one finger; two fingers and three fingers were evaluated in \cite{Al-Nima2017finger}. 

A novel combination method termed the Re-enforced Probabilistic Neural Network (RPNN) was suggested. This approach was applied to verifying people by utilising two FT sets. That is, two FT databases that were acquired under different spectra from the CASIAMS were used. The RPNN examines the first FT inputs, then, it can exploit the second FT inputs to confirm the verification. The combination between the FT sets is performed in the decision level. It provided high recognition performance by using the FTs of only four fingers \cite{al2018personal}. Nevertheless, it requires two sets of FTs from each user. 

Various fusion methods were evaluated to combine FTs of the CASIAMS (Spectral 460) database and CASIAMS (Spectral White) database. Two sets of the main four fingers were used for each person. Four	levels of fusion were examined: sensor level; feature level; score level and decision level. Furthermore, the following rules were considered: average; summation; multiplication; maximum; minimum and concatenation. The concatenation rule was not applied to the decision level fusion because that is not feasible, therefore, the AND and OR rules were applied instead \cite{al2018finger}. 

A summary of multi-object fusions that employed the FT characteristic are given in Table \ref{Table:Fusion}. As it can be observed, the FTs of fingers were not always fully employed and were usually combined with other biometrics such as the palmprint. One can argue that the fingers of a hand can contribute together to give precise recognition decision. Therefore, a multi-object biometric system based on only the FTs of finger objects can always be considered. In this case, a single acquirement equipment and a single feature extraction method can be used for all the FTs, this will reduce the cost of providing an additional acquiring device and establishing an extra feature extraction algorithm.  
\begin{table}[!h]
	\centering
	\caption{Details of multi-object fusions that were employed in FT characteristic studies}
	\label{Table:Fusion}
	\scalebox{0.65}{\begin{tabular}{|C{4cm}|C{5cm}|C{2.5cm}|C{6cm}|}
			\hline
			\textbf{Reference} &  \textbf{Employed objects} & \textbf{Fusion level} & \textbf{Fusion rule} \\ \hline
			Ribaric and Fratric \cite{Ribaric2005Anonline}		&  Four FTs and fingers-geometry & Score level & Weighted summation \\ \hline
			Ribaric and Fratric \cite{Ribaric2005ABiometric}		&  Five FTs and palmprint & Score level & Weighted summation \\ \hline
			\multirow{2}{4cm}{\centering Ying \textit{et al.} \cite{ying2007identity}} 		& \multirow{2}{4cm}{\centering Five FTs and palmprint} & Feature level  & Holistic\\ \cline{3-4}
			& & Score level & Weighted summation\\ \hline
			\multirow{2}{4cm}{\centering Nanni and Lumini \cite{nanni2009multi}} 		& Middle finger & Score level & Summation \\ \cline{2-4}
			& Middle finger and ring finger	&  Score level & Summation\\ \hline
			Pavesic \textit{et al.} \cite{Pavesic2009Finger-based} 	& Four FTs and four fingerprints &  Score level & Weighted summation\\ \hline
			Michael \textit{et al.} \cite{Michael2010Robust}		& FTs and palmprint & Score level & SVM \\ \hline
			Michael \textit{et al.} \cite{michael2010innovative} 		& FTs and palmprint & Score level & SVM \\ \hline
			\multirow{4}{4cm}{\centering Goh \textit{et al.} \cite{Goh2010Bi-modal}} 		& \multirow{4}{4cm}{\centering FTs and palmprint} & \multirow{4}{2cm}{\centering Score level} & AND \\ \cline{4-4}
			& & & OR \\ \cline{4-4}
			& & & Summation\\ \cline{4-4}
			& & & Weighted summation\\ \hline
			Zhang \textit{et al.} \cite{Zhang2010hand}  		& Middle finger and palmprint & Feature level & Concatenation \\ \hline
			\multirow{4}{4cm}{\centering Kanhangad \textit{et al.} \cite{Kanhangad2011AUnified}} 		& FTs, 2D palmprint and 2D hand geometry 
			& Score level & Weighted summation\\ \cline{2-4}
			& FTs, 2D palmprint, 2D hand geometry, 3D palmprint and 3D hand geometry & Score level & Weighted summation\\ \hline
			\multirow{2}{4cm}{\centering Zhang \textit{et al.} \cite{zhang2012hand}} & \multirow{2}{4cm}{\centering Middle finger and palmprint} & Feature level  & Weighted concatenation\\ \cline{3-4}
			& 	& Score level & Summation\\ \hline
			\multirow{8}{4cm}{\centering A. Kumar and Y. Zhou \cite{Kumar2012Human}} 		& \multirow{2}{4cm}{\centering Part of index finger and index finger veins} & \multirow{2}{2cm}{\centering Score level} & holistic \\ \cline{4-4}
			& & &non-linear\\ \cline{2-4}
			& \multirow{2}{4cm}{\centering Part middle finger and middle finger veins} & \multirow{2}{2cm}{\centering Score level} & holistic \\ \cline{4-4}
			& & & non-linear\\ \cline{2-4}
			& \multirow{4}{3.5cm}{\centering Part of index finger, part of middle finger, index finger veins and middle finger veins} & \multirow{4}{2cm}{\centering Score level} & \multirow{2}{3cm}{\centering holistic} \\ 
			& & & \\ \cline{4-4}
			& & & \multirow{2}{3cm}{\centering non-linear}\\ 
			& & & \\ \hline
			Al-Nima \textit{et al.} \cite{Al-Nima2015Human} 		& Four FTs & Feature level & Concatenation \\ \hline
			Al-Nima \textit{et al.} \cite{Al-Nima2016ANovel} 		& Four FTs & Feature level & Concatenation \\ \hline
			Al-Nima \textit{et al.} \cite{Al-Nima2017Robust} 		& Five FTs & Feature level & Concatenation \\ \hline
			Al-Nima \textit{et al.} \cite{Al-Nima2017efficient} 	& Five FTs & Feature level & Concatenation \\ \hline
			\multirow{2}{*}{Al-Nima \textit{et al.} \cite{Al-Nima2017finger}} 		& Five FTs & Feature level & Concatenation \\ \cline{2-4}
			& Five FTs & Score level & Summation \\ \hline
			
			\multirow{4}{*}{MAC \textit{et al.} \cite{MAC2018contactless}} & \multirow{3}{*}{Three segments from index finger} & Feature level & Concatenation\\ \cline{3-4}
			& & Sensor level & Concatenation \\ \cline{3-4}
			& & Sensor and Feature levels & Concatenation \\\hline
			
			MAC \textit{et al.} \cite{Jahan2018Contactless} & Three segments from index finger & Sensor level & Concatenation \\\hline
			%			Chopra \textit{et al.} \cite{Chopra2018Unconstrained} & Index and middle fingers & Sensor level & Concatenation \\\hline
			
			Al-Nima \textit{et al.} \cite{al2018personal} & Four FTs & Decision level & RPNN \\ \hline
			
			\multirow{9}{*}{Al-Kaltakchi \textit{et al.} \cite{al2018finger}} & Four FTs & Sensor level & Average; summation; multiplication;	maximum; minimum and concatenation\\ \cline{2-4}
			& Four FTs & Feature level & Average; summation; multiplication;	maximum; minimum and concatenation \\ \cline{2-4}
			& Four FTs & Score level & Average; summation; multiplication;	maximum; minimum and concatenation \\ \cline{2-4}
			& Four FTs & Decision level & Average; summation; multiplication;	maximum; minimum; AND and OR \\ \hline
			
	\end{tabular}}
\end{table}

\section{FT recognition performances}
To evaluate the FT recognition performances, a number of common measurements were used. These measurements can be illustrated as follows:

$\bullet$ False Acceptance Rate (FAR): also known as False Match Rate (FMR) or False Positive Rate (FPR)\cite{%Woods1997Generating,
	saugirouglu2009Intelligent}. It is the ratio between number of accepted imposters to the total number of imposters \cite{Goh2010Bi-modal}. 
%	It can be represented by the following equation \cite{Goh2010Bi-modal}:\\
%	\begin{equation}
%		FAR = \frac{Number~of~Accepted~Imposters}{Total~Number~of~Imposters} \times 100\%
%	\end{equation}\\

$\bullet$ False Rejection Rate (FRR): also known as False Non-Match Rate (FNMR) %\cite{Woods1997Generating} 
\cite{saugirouglu2009Intelligent}. It is the ratio between the number of rejected clients to the total number of clients \cite{Goh2010Bi-modal}. 
%	It can be represented by the following equation \cite{Goh2010Bi-modal}:\\
%	\begin{equation}
%		FRR = \frac{Number~of~Rejected~Clients}{Total~Number~of~Clients} \times 100\%
%	\end{equation}\\

$\bullet$ Equal Error Rate (EER): It is the trade off point between the FAR and FRR. It is considered as an essential parameter to evaluate any biometric system rather than the FT recognition system. Basically, if the EER has a small value this means that the system is efficient and vice versa. Statistically, the EER is equivalent to a value of threshold in which the FRR = FAR \cite{meshoul2010novel}.

$\bullet$ True Positive Rate (TPR): Also known as Genuine Acceptance Rate (GAR) \cite{Woods1997Generating} \cite{saugirouglu2009Intelligent} or True Acceptance Rate (TAR) 
%		\cite{Debayan2018matching}
. It is the ratio between the number of correctly classified positives to the total number of clients. Mathematically, It equals to 1-FRR. 

$\bullet$ Recognition rate (or recognition accuracy): It represents the recognition accuracy of a biometric system \cite{SAZAK2019TheMultiscale}. Effectively, it determines how the system is successful in a percentage value.

$\bullet$ Receiver Operating Characteristic (ROC): It is a curve that represents the relationship between the FAR and TPR (or 1-FRR) \cite{WILD2016Robust}. The ROC is widely used	to report the recognition system measurements.

$\bullet$ Area Under the Curve (AUC): It is a value that shows the area under the ROC curve. In other words, it measures the occupied area by the ROC cure.

$\bullet$ Detection Error Tradeoff (DET): It is a curve that represents the relationship between the FAR and FRR.

$\bullet$ Cumulative Match Characteristics (CMC): It is a curve that represents the relationship between the recognition rate (or recognition accuracy) and cumulative rank. This curve is employed to show the identification performance \cite{Kumar2012Human}.

$\bullet$ Time: It usually represents an average time of implementing recognition operation(s) \cite{jaswal2016knuckle}. This evaluation has no standard equipments to be measures as it depends on the specifications of the used equipment parts during the operations. So, usually the specifications of the exploited recognition device is described for this measurement.
%It has been illustrated that reducing the FAR value will increase the security level of the system, whilst, increasing the FAR value will increase the flexibility of the system in terms of accepting input subjects \cite{maltoni2009handbook}.
The recorded FT performances are clarified as follows: Ribaric and Fratric \cite{Ribaric2005Anonline} established the first combination system between the FTs and fingers-geometry, a scanner device was used to acquire the collected hand images. 
%Both the identification and verification experiments were utilised in this study. 
The best FT performances were recorded as 3.51\% for the identification and 0.26\% for the verification. After combining the FT with the geometry of fingers the EER for the identification declined to 1.17\% and for the verification reduced to 0.04\%. 

The same authors designed a combination system between the FTs and palmprint \cite{Ribaric2005ABiometric}. The EER of only the FTs was not reported. However, the overall performances were reported, where two identification experiments were executed. The early experiment attained the lowest EER of 0.58\%. The low cost fusion biometric system between the palm, hand geometry and FTs that was designed by \cite{Ferrer2007Low} achieved 0.13\% for the EER of using only the FTs. Different fusion methods were applied in the case of identification recognition. The best recognition performance after the combination of all the exploited characteristics was obtained by the decision fusion, where the FAR was equal to 0\% and the FRR was equal to 0.15\%.
%The hand images were captured by an available scanner with a resolution of 150 dpi. 
%A simple contour was utilised after the image binarization by using the Otsu threshold \cite{otsu1975threshold}. Then, the contour cartesian coordinates were converted to polar coordinates in order to calculate the locations of the tip and valley points of the fingers from the middle point of the hand base, where a hand was located in a specific location. Noticeably, not all patterns of the FTs were applied and a fixed location was used to acquire the hand images which did not allow free translations or orientation movements to be examined. 

The main achievement of the \cite{ying2007identity} study was addressing the effects of different hand poses. It could segment the palmprint and the five fingers of normally stretched hand parts (completed hand area should be included in the image). There was no benchmarked recognition values and no improvement fusion recognition rate. 
%and failed with the others. This is due to how successfully the hand contour is established. 
%"This paper presents and compares two hand texture based personal identification methods, which are called hand-print verification in this paper to denote the idea of utilising whole hand skin image for recognition."
Only one or two fingers were considered in \cite{nanni2009multi}, where low resolution images acquired by a camera were utilised. By employing both the middle and ring fingers the EER value was between 0.18\% and 0\% according to the parameters of the proposed BioHashing multi-matcher. The problem here is that the fingerprint region was included, and its features were not considered. So, it seems that a wasting area was involved with the FT.

%"Finally, considering both middle and ring fingers allows the performance to reach a near zero EER; the multimatcher MM (BioHashing) described above obtains an EER of zero in the BEST hypothesis and an EER of 0.18 in the WORST hypotheses in the tested dataset."
In \cite{Pavesic2009Finger-based}, the authors also used a scanner acquisition device to collect part of hand images. So, they used high resolution parts of hand images located in a limited space. As mentioned, the fingerprints were fused with limited areas of FTs for the four fingers of each subject. The best identification recognition rate was 99.98\% and the best verification EER value was 0.01\% after the fusion.

Combinations between the FTs and palmprint were presented in \cite{Michael2010Robust}, \cite{michael2010innovative} and \cite{Goh2010Bi-modal}. In these publications, a Charge-Coupled Device (CCD) camera was used to collect video streams of hand images. Different verification specifications were used, as clarified in the table. The EERs of only the FTs were reported to be 4.10\%, 1.95\% and 2.99\% for \cite{Michael2010Robust}, \cite{michael2010innovative} and \cite{Goh2010Bi-modal}, respectively. These values were enhanced after the combinations to the EER values of 0.0034\% and 1.25\% for \cite{Michael2010Robust} and \cite{Goh2010Bi-modal}, respectively, and to the recognition rate of  99.84\% for \cite{michael2010innovative}.
A framework study utilising the FTs as a part of fusion between palmprint, hand geometry and finger surfaces from 2D and 3D hand images to enhance the contactless hand verification was applied by \cite{Kanhangad2011AUnified}. High EER value equal to 6\% was benchmarked for the FTs and this percentage declined to 0.22\% after the fusion between all the utilised characteristics.

%From Table \ref{table:four_five_fingers3} it is clear that using more features will increase the successful performance of the verification. 
Kumar and Zhou \cite{kumar2011contactless} illustrated a biometric identification method by using a very small part part of the FT with a part of a fingerprint. The Hong Kong Polytechnic University Low Resolution Fingerprint (PolyULRF) Database (Version 1.0) database database was exploited. The EER here reached 0.32\%. Similarly, A. Kumar and Y. Zhou \cite{Kumar2012Human} explained extensive identification work by employing the same database of PolyULRF to combine between a very small part of the finger surface with the finger vein. The EER results of small part of FTs with parts of the fingerprints were as follows: for the index finger 0.32\%, for the middle 0.22\% and for both the index and middle fingers 0.27\%. The best EER was obtained after the combination but by using only the middle finger, where it was equal to 0.02\%.

The main idea of \cite{zhang2012hand} is to utilise the features of the middle finger and the palmprint in a Single Sample Biometrics Recognition (SSBR), where both can be acquired by using a single hand image sample. Recognition rates were used in this study instead of the EER to show the recognition performance. The number of participants were 100, each individual provided 10 images. So, a total of 1000 images was used. This collected database was partitioned into 100 images, first image from each subject, to be stored in a template and 900 images to be assessed. 
%The best identification performance was the one that achieved the highest recognition rate, whereas, the best verification performance was the one that achieved the lowest EER value. 
In terms of identification: the recognition rate attained by using the feature level fusion for the middle finger was 98.33\%; the recognition rate achieved by utilising the feature level fusion for the palmprint was 95.78\% and the recognition rate obtained by applying the score fusion to both the FLF, was increased to 99.56\%. In terms of verification: the EER value attained by using the feature level fusion for the middle finger was 1.09\%; the EER value achieved by utilising the feature level fusion for the palmprint was 1.98\% and the EER value obtained by applying the score fusion to both the feature level fusions was decreased to 0.49\%. Again, areas of fingerprints were wasted here without extracting their specific features. 

%	In \cite{stein2013video}, the determined ``outer area", where a part of the FT is included, was basically applied for the negative authentication. Principally, in the negative authentication the individual request is only tested for the recognition validity \cite{dasgupta2017negative}. So, there was no recognition performance assigned for the FT. The used database images were collected by ``Galaxy Nexus" and ``Nexus S" smartphones from Samsung. The tested photos were captured as: 541 and 569	images the ``Galaxy Nexus" and ``Nexus S", respectively. Also, 990 images were acquired by the "Galaxy Nexus" smartphone as videos. Overall, a total of 2100 tested finger samples were considered. 

As mentioned, Bhaskar and Veluchamy \cite{Bhaskar2014Hand} suggested a multi-modal biometric verification system based on feature fusion between the FTs and palmprints. This study used the IIT Delhi palmprint database, but did not describe the partitioning of training and testing sets. After the combination, the recognition rate attained 98.5\%.

%	A verification approach was suggested in \cite{sankaran2015Onsmartphone}, where the IIITD fingerphoto database was reported. The main observation in this database is that it was collected under different illumination and background variations, white indoor/outdoor backgrounds and natural indoor/outdoor backgrounds. The database was randomly divided into 50\% images for gallery and 50\% images as probes for the test. By using the white indoor images in the gallery, natural outdoor images achieved the best EER value of 3.65\%. The problem here is that the fingerprint was employed with a part of the FT. 

An important FT study was introduced by \cite{Al-Nima2015Human}. In this publication, the FT region was assigned and all the FT parts were determined. It confirmed that using more FT features increase the successful performance of the verification. The EERs after adding the third or lower knuckle were better than the EERs without this important part. This issue was recorded in different feature extraction methods such as in the IFE based exponential histogram the EER percentage was reduced from 5.42\% to 4.07\% and in the IFE based bell-shaped histogram the EER value declined from 12.66\% to 7.01\%. 

The work in \cite{Al-Nima2016ANovel} was mainly established to produce a novel approach of generating the ROC graph from the PNN, as mentioned. Therefore, enhancing the recognition performance was not essential. A well-known feature extraction called the LBP obtained the best performance in that paper with an EER equal to 1.81\%.

%	The recognition specifications in \cite{malhotra2017fingerphoto} is very similar to \cite{sankaran2015Onsmartphone} as the same database, result and problem were considered. 
In \cite{Al-Nima2017Robust}, a robust finger segmentation method, efficient feature extraction method to extract the main FT patterns and a novel salvage approach to rescue the missing FT features were illustrated. It was applied for all the five fingers and attained results consistent with \cite{Al-Nima2015Human}, where by adding the FTs of the thumb the verification performances were enhanced. The best EER values for the three databases the Hong Kong Polytechnic University Contact-free 3D/2D (PolyU3D2D), IIT Delhi and CASIA Multi-Spectral (CASIAMS) (Spectral 460) were equal to 0.11\%, 1.35\% and 3\%, respectively. It is worth mentioning that the proposed salvage approach for the missing FT parts has the capability to enhance the verification performance. That is when an amputation may happen to the employed fingers, the salvage approach can be used with the PNN to reduce the risk of obtaining a wrong verification. 

Efficient finger segmentation approach was suggested in \cite{Al-Nima2017efficient}, where it was established for this reason. This paper exploited the LLBP as a feature extraction. This method achieved reasonable performances, due to the fact that it considers the vertical and horizontal lines in its operator and this is appropriate for the main patterns of the FTs. Since the best lengths of the LLBP vectors are (N=13, N=15, N=17 and N=19) as suggested in \cite{Petpon2009Face}, all of these lengths were considered. The best performances were obtained by the lengths N=13 (and 17), N=13, and N=19 for the PolyU3D2D, IIT Delhi and CASIAMS (Spectral 460) databases, respectively. For the same order, the best EER percentages were 0.68\%, 2.03\% and 5\%, respectively.

The FCFNN that proposed in \cite{Al-Nima2017finger} was specifically designed to fuse the FTs of fingers. This approach was inspired from the contribution of each finger. For instance, the contribution of the thumb finger is not equal to the contribution of the index or middle finger. The recognition performance was enhanced for the PolyU3D2D database from 0.68\% to 0.23\% after using the FCFNN with the MSALBP feature extraction. Also, the best EER value was 2\% for the CASIAMS (Spectral 460) database after using the PNN with the MSALBP and after using the FCFNN with the MSALBP too. In this study the effects of the amputated fingers were also considered by taking advantages from the flexible architecture of the FCFNN. 

%	Debayan \textit{et al.} \cite{Debayan2018matching} collected fingerphoto and fingerprint images in India from 309 subjects. The fingerphoto images were obtained by using Xiaomi Redmi Note 4 smartphone. Fingerprints with small parts of FTs were considered in this study for two index fingers and two thumb fingers from each hand. Furthermore, two mobile phone applications were exploited in the case of verification. Total of 4,944 genuine scores (309	subjects $\times$ 4 fingers $\times$ 2 enrollment impressions $\times$ 2 verification impressions) and total of 95,172 impostor scores (309 fingers $\times$ 308 fingers) were evaluated. The performances were generally reported by calculating the TAR and fixing the FAR value to 0.1\%. The TAR was equal to 72.14\% in the first application and 99.66\% in the second application for the fingerphoto-to-fingerphoto matching and for fusing between all four employed fingers.

MAC \textit{et al.} \cite{MAC2018contactless} presented a verification study based on contactless multiple finger segments. Finger images acquired by a 24 Megapixel digital camera from a number of video frames (around 20-40) that covers 1.5 second interval for each participant. The participants were from different nationalities and they were asked to provide various rotation and scaling movements. 1341 images were utilised for 41 individuals. The authors concentrated on contactless multiple index finger segments, where three areas were segmented from the three finger knuckles. Various types of fusions based on the concatenation rule were considered between the finger segments feature level; sensor level; and feature and sensor levels. The recognition (classification) accuracies for the three combination methods were respectively recorded as follows 95.37\%; 97.70\% and 97.93\%. Obviously, the best performance was obtained by using the feature and sensor levels fusion method. 10-fold cross validation and SVM (RBF kernel) were used to evaluate the performance. 

MAC \textit{et al.} \cite{Jahan2018Contactless} exploited the same facilities and tools of the previous work. Nonetheless, only the sensor level combination with concatenation rule was used and the recognition (classification) accuracy of  92.99\% was calculated with a negative group of 49889 samples and a positive group of 1238 samples.

Omar \textit{et al.} \cite{omar2018deep} attained interesting results in terms of verification, where the deep learning is exploited with the FT and a novel DFTL has been approached. Four databases were employed in this study. These are PolyU3D2D; IIT Delhi; CASIAMS (Spectral 460) and CASIAMS (Spectral White), recognition accuracies of 100\%; 98.65\%; 100\% and 98\% were respectively achieved. The drawback of this work is considering any successful FT in any five fingers to confirm the verification.

Al-Nima \textit{et al.} \cite{al2018personal} suggested two new methods. One for the feature extraction called the SPC and another for the combination named the RPNN. Two databases of FT that acquired under two different lighting from the CASIAMS is utilised in this work. These are the CASIAMS (Spectral 460) and CASIAMS (Spectral White). FTs of only the four or main fingers were applied. The EER values of using the SPC feature extraction with normal PNN 
%	\cite{PNN_MatlabCode} 
were equal to 4\% and 2\% for the CASIAMS (Spectral 460) and CASIAMS (Spectral White) databases, respectively. High verification performances of EER equal to 0\% were obtained after using the novel RPNN. 

Al-Kaltakchi \textit{et al.} \cite{al2018finger} examined different fusion levels between FTs databases of the CASIAMS (Spectral 460) and CASIAMS (Spectral White). Two sets of four fingers were utilised for each person in this study too. The four	fusion levels were evaluated: sensor level; feature level; score level and decision level. Various rules were considered: average; summation; multiplication;	maximum; minimum and concatenation. Only the concatenation rule could not be used with the decision level as this is not applicable, AND and OR rules were employed instead. Best recognition performance was reported for the	feature level fusion with the concatenation rule as the EER reached its lowest value of 2\%.
\begin{table}[!b]
	\centering\scriptsize
	\caption{Best FT performances for the presented recognition work with their specifications}
	\label{Table:FT_performances}
	\scalebox{0.65}{\begin{tabular}{|C{2cm}|C{2.3cm}|C{2.5cm}|C{2.6cm}|C{2cm}|C{2.5cm}|C{2cm}|}
			\hline
			\textbf{Reference} & \textbf{FT database type} & \textbf{Acquisition device} & \textbf{Number of employed subjects} & \textbf{Recognition type} & \textbf{Number of tested FTs\footnote{This table has been derived from the number of tested image samples per each participant multiplied by the number of used fingers.}} & \textbf{Best EER value (\%)} \\ \hline
			\multirow{3}{2cm}{\centering Ribaric and Fratric \cite{Ribaric2005Anonline}}	 & \multirow{3}{2cm}{\centering Collected images} & \multirow{3}{2cm}{\centering A scanner (180 dpi)} & \multirow{3}{1cm}{\centering 127} &  Identification & 684  clients / 2800 impostors & 3.51 \\ \cline{5-7}
			& & & & Verification & 684 clients / 159600 impostors & 0.26 \\ \hline
			Ribaric and Fratric \cite{Ribaric2005ABiometric} & Collected images & A scanner (180 dpi) & 237 & Identification & 855 clients / 3500 impostors & --- \\ \hline
			Ferrer \textit{et al.} \cite{Ferrer2007Low} & Collected images & A scanner (150 dpi) & 109 & Identification & 2616 clients / 282528 impostors & --- \\\hline
			
			Ying \textit{et al.} \cite{ying2007identity}   & UST (Not available) & A camera (150 dpi) & 287 & Identification & 129150 clients / 7400295 impostors & --- \\ \hline
			Nanni and Lumini \cite{nanni2009multi} & Collected images & A camera & 72 & Verification & 720 & --- \\ \hline
			\multirow{3}{2cm}{\centering Pavesic \textit{et al.} \cite{Pavesic2009Finger-based}} & \multirow{3}{2cm}{\centering Collected images} & \multirow{3}{2cm}{\centering A scanner (600 dpi)} & \multirow{3}{1cm}{\centering 184} & Identification & 3680 clients / 0 impostors & --- \\ \cline{5-7}
			& & & & Verification & 2760 clients / 253920 impostors & --- \\ \hline
			Michael \textit{et al.} \cite{Michael2010Robust} & Collected video stream & CCD web camera & 50 & Verification & 10-cross validation of 2500 & 4.10 \\ \hline
			Michael \textit{et al.} \cite{michael2010innovative} & Collected video stream  & CCD web camera & 100 & Verification & 18000 clients / 198000 imposters & 1.95 \\ \hline
			Goh \textit{et al.} \cite{Goh2010Bi-modal} & Collected video stream & CCD web camera & 125 & Verification & 22500 clients / 315000 imposters & 2.99 \\ \hline
			Zhang \textit{et al.} \cite{Zhang2010hand} & Collected images & CCD camera & 98 & Identification & 11-cross validation of 980 & --- \\ \hline
			Kanhangad \textit{et al.} \cite{Kanhangad2011AUnified} & PolyU3D2D & Minolta VIVID 910 & 177 & Verification & 3540 & 6 \\ \hline
			Kumar and Zhou \cite{kumar2011contactless} & PolyULRF & Web camera & 156 & Identification & 936 clients / 145,080 imposters & --- \\ \hline
			A. Kumar and Y. Zhou \cite{Kumar2012Human} & PolyULRF & Web camera & 156 & Identification & 936 clients / 145,080 imposters & --- \\ \hline
			\multirow{2}{2cm}{\centering Zhang \textit{et al.} \cite{zhang2012hand}} & \multirow{2}{2cm}{\centering Collected images} & \multirow{2}{2cm}{\centering A camera} & \multirow{2}{1cm}{\centering 100} & Identification & 900 & --- \\ \cline{5-7}
			&  &  &  & Verification & 900 & --- \\ \hline
			%			Stein \textit{et al.} \cite{stein2013video} & Collected images / videos & Smartphone camera & 37 & Verification & 2100 & --- \\ \hline
			Bhaskar and Veluchamy \cite{Bhaskar2014Hand} & IIT Delhi & A camera & Not given & Identification & Not given & --- \\ \hline
			%			Sankaran \textit{et al.} \cite{sankaran2015Onsmartphone} & IIITD & Smartphone camera & 128 & Verification & 2048 & --- \\ \hline
			Al-Nima \textit{et al.} \cite{Al-Nima2015Human} & PolyU3D2D & Minolta VIVID 910 & 177 & Verification & 3540 & 4.07 \\ \hline
			Al-Nima \textit{et al.} \cite{Al-Nima2016ANovel} & PolyU3D2D & Minolta VIVID 910 & 177 & Verification & 3540 & 1.81 \\ \hline
			%			Malhotra \textit{et al.} \cite{malhotra2017fingerphoto} & IIITD & Smartphone camera & 128 & Verification & 2048 & --- \\ \hline
			\multirow{4}{2cm}{\centering Al-Nima \textit{et al.} \cite{Al-Nima2017Robust}} & PolyU3D2D & Minolta VIVID 910 & 177 & Verification & 4425 & 0.34 \\ 	\cline{2-7}
			&  IIT Delhi & A camera & 148 & Verification & 740 & 1.35 \\ \cline{2-7}
			&  CASIAMS (Spectral 460nm) & CCD camera & 100 & Verification & 500 & 3 \\ \hline
			\multirow{4}{2cm}{\centering Al-Nima \textit{et al.} \cite{Al-Nima2017efficient}} & PolyU3D2D & Minolta VIVID 910 & 177 & Verification & 4425 & 0.68 \\ \cline{2-7}
			&  IIT Delhi & A camera & 148 & Verification & 740 & 2.03 \\ \cline{2-7}
			&  CASIAMS (Spectral 460nm) & CCD camera & 100 & Verification & 500 & 5 \\ \hline
			\multirow{3}{2cm}{\centering Al-Nima \textit{et al.} \cite{Al-Nima2017finger}} & PolyU3D2D & Minolta VIVID 910 & 177 & Verification & 4425 & 0.23 \\ \cline{2-7}
			&  CASIAMS (Spectral 460nm) & CCD camera & 100 & Verification & 500 & 2 \\ \hline
			%			Debayan \textit{et al.} \cite{Debayan2018matching} & Collected images & Xiaomi Redmi Note 4 smartphone & 309 & Verification & 4,944 clients / 95,172 impostor & --- \\ \hline
			MAC \textit{et al.} \cite{MAC2018contactless} & Collected video frames & 24 Megapixel digital camera & 41 & Verification & 10-fold cross validation & --- \\ \hline
			
			MAC \textit{et al.} \cite{Jahan2018Contactless} & Collected video frames & 24 Megapixel digital camera & 41 & Verification & 10-fold cross validation & --- \\ \hline
			
			%			Wasnik \textit{et al.} \cite{Wasnik2018Improved} & Collected video frames & iPhone 6s & 48 & Verification & 240 & --- \\ \hline
			
			%			Wasnik \textit{et al.} \cite{wasnik2018baseline} & Collected video frames & iPhone 6s & 48 & Verification & 240 clients / 11280 imposter & --- \\ \hline 
			
			%			\multirow{3}{2cm}{\centering Weissenfeld \textit{et al.} \cite{Weissenfeld2018contactless}} & \multirow{3}{2cm}{\centering Collected images} & LG G5 850 mobile phone and Huawei P9 mobile phone & 12 & Verification & 1920 images & --- \\ \cline{3-7}
			%			&   & Handheld embedded device & 94 & Verification & Not clear & --- \\ \hline
			%			Chopra \textit{et al.} \cite{Chopra2018Unconstrained} & Different smartphones such as iPhone; Google Nexus and Samsung Galaxy & Smartphone camera & 115 & Verification & Not clear & --- \\ \hline

			\multirow{6}{2cm}{\centering Omar \textit{et al.} \cite{omar2018deep}} & PolyU3D2D & Minolta VIVID 910 & 177 & Verification & 4425 & --- \\ \cline{2-7}
			&  IIT Delhi & A camera & 148 & Verification & 740 & --- \\ \cline{2-7}
			&  CASIAMS (Spectral 460nm) & CCD camera & 100 & Verification & 500 & --- \\ \cline{2-7}
			&  CASIAMS (Spectral White) & CCD camera & 100 & Verification & 500 & --- \\ \hline
			\multirow{3}{2cm}{\centering Al-Nima \textit{et al.}
				\cite{al2018personal}} 			&  CASIAMS (Spectral 460nm) & \multirow{3}{2cm}{\centering CCD camera} & \multirow{3}{2cm}{\centering 100} & \multirow{3}{2cm}{\centering Verification} & \multirow{3}{2cm}{\centering 800} & \multirow{3}{2cm}{\centering 0} \\ \cline{2-2}
			& CASIAMS (Spectral White) & & & & & \\ \hline 
			\multirow{3}{2cm}{\centering Al-Kaltakchi \textit{et al.}}&  CASIAMS (Spectral 460nm) & \multirow{3}{2cm}{\centering CCD camera} & \multirow{3}{2cm}{\centering 100} & \multirow{3}{2cm}{\centering Verification} & \multirow{3}{2cm}{\centering 800} & \multirow{3}{2cm}{\centering 2} \\ \cline{2-2}
			& CASIAMS (Spectral White) & & & & & \\ \hline

	\end{tabular}}
\end{table}

Best EER performances of the presented FT work with their specifications are given in Table \ref{Table:FT_performances}. As it can be seen the performances of the suggested FT recognition approaches are generally still requiring more improvements. Large numbers of FT samples can be easily acquired. Many effective ideas can be proposed to perform FT biometric systems with high levels of accuracy. Furthermore, many approaches can be adopted to enlarge the research areas of the FT recognition field. 

\section{Conclusion and Future Work}
The findings of this survey have found a number of important observations in FT studies. Firstly, for the finger segmentation and ROI collection, the majority of the previous studies considered limited areas of the FT regions. Secondly, FT patterns lacked a beneficial feature extraction model that can efficiently collect its ridges, visible lines and skin wrinkles features. Thirdly, usually the FTs were combined with supported biometric characteristic(s) to construct multi-modal biometric structures. Whereas, this trait can be used alone as multi-objects by considering the FT of each finger as a single object. 

%	The employed and available FT databases were reviewed. These are as follows: the PolyU3D2D; the IIT Delhi Database and the CASIAMS (Spectral 460), PolyUFI and PolyULRF%IIITD databases. 
Also, essential observations have been notices. The databases which have been used are fundamentally established for palmprint or fingerprint studies. So, it is believed that a specific FT database is required, where this database has to cover the full FT regions and involve all of the feature types (wrinkles, visible lines and ridges).
%From the previous literature, it can be observed that the FTs have been used as a part of a multi-modal biometric scheme. In addition, limited FT features have been utilised as the majority of the ROI extracting approaches collected just part of the FTs. 
%It can be argued that reliable biometric recognition approaches can be established by exploiting the FT patterns of multiple fingers in a fused multi-object prototype. This will reduce the cost of providing an additional acquiring device and establishing an extra applicable algorithm for including another biometric trait to be combined with the FTs. \\
%In addition, many publications completely ignored the minutiae patterns.

Regarding the recognition performances, it can be concluded that the FTs were efficiently exploited in many recognition systems. A large number of FT samples is easy to be collected. The FTs can provide effective recognition contributions if they are exploited in single-modal or multi-modal biometric systems. Nevertheless, the recognition performance based on the FT(s) can be further improved. Also, many ideas can be adopted to increase the investigations of this field. 

Moreover, the following insights are suggested for future work:

$\bullet$ Biometric application projects and systems based on the FT(s) can be produced. That is, commercial security systems and applications are easily designed and implemented. Moreover, multi-object of FTs can effectively be exploited.

$\bullet$ There is no sufficient information for the ridge patterns of the FT. So, they require to be focused upon in future studies. In this case, a high resolution camera or scanner is required to be employed. The ridge information can also be obtained by using smartphones
%just as the fingerphoto images
. Android applications may be employed for this matter.

$\bullet$ The permanency of the ridge patterns needs to be examined in order to see if they are also vanished for the elderly people like the fingerprint or not. Thus, a long term of observations is important to be considered for the FT ridge patterns. This might be reported as a novel approach.

$\bullet$ A comprehensive database is required. It should include all the FT patterns (wrinkles, visible lines and ridges). This database has to concentrate on the FT region only. Different rotations, translations and scaling are worth to be examined. 

$\bullet$ In the case of finger segmentations, additional efforts will be required to generate a robust ground truth. The current suggested ground truth is based on the essential points of fingers (tips and valleys). Obviously, these points can not cover all the patterns of the lower knuckles. So, a justified ground truth for the finger segmentation can be established then provided to other researchers. 

$\bullet$ Injured or uncleared FT patterns are worth to be investigated in terms of recognition performances. Moreover, The affects of skin diseases are also need to be considered in future studies. Some of these diseases can be overcomes by the power of image processing such as the rash. Others may influence the pattern shapes such as acnes. Nevertheless, the wide area of spreading FT patterns may help the biometric systems to stay obtaining the correct recognition.

$\bullet$ The terminology of Finger Inner Surface (FIS) can be used in future work as it provides the same meaning of FT. In addition, an intensive study may be established to decide which part of the FT is more effective in terms of individual recognition. Another study to compare between the FTs of males and females needs to be introduced.

$\bullet$ Hyperspectral imaging of FTs are worth studying, where interesting FT patterns and textures are revealed according to the afforded electromagnetic spectra. Furthermore, multi-spectral sensors can be used to reveal different FT patterns. Then, fusion studies can be performed between the different extracted patterns.

\section*{Acknowledgement}
\begin{small} This work is supported by EPSRC grants (EP/P015387/1, EP/P00430X/1);  Birkbeck BEI School
	Project (ARTEFACT); Guangdong Science and Technology Department grant (No. 2018B010107004), China. 
\end{small}

%	$\bullet$ "The  Contact-free 3D/2D Hand Images Database version 1.0". \\
%	$\bullet$ "IIT Delhi Palmprint Image Database version 1.0".\\
%	$\bullet$ "Portions of the research in this paper use the CASIA-MS-PalmprintV1 collected by the Chinese Academy of Sciences' Institute of Automation (CASIA) ".

%\section*{Bibliography}
% Bibliography
%\begin{small}
\bibliographystyle{elsarticle-num}
\bibliography{Finger_Texture_Biometric_Characteristic_A_Survey}

\begin{thebibliography}{10}
\expandafter\ifx\csname url\endcsname\relax
  \def\url#1{\texttt{#1}}\fi
\expandafter\ifx\csname urlprefix\endcsname\relax\def\urlprefix{URL }\fi
\expandafter\ifx\csname href\endcsname\relax
  \def\href#1#2{#2} \def\path#1{#1}\fi

\bibitem{li2004personal}
Q.~Li, Z.~Qiu, D.~Sun, J.~Wu, Personal identification using knuckleprint, in:
  Advances in Biometric Person Authentication, Springer, 2004, pp. 680--689.

\bibitem{Al-Nima2017Signal}
R.~R.~O. Al-Nima, Signal processing and machine learning techniques for human
  verification based on finger textures, {PhD} thesis, School of Engineering,
  Newcastle University (2017).

\bibitem{al2019segmenting}
R.~R.~O. Al-Nima, N.~A. Al-Obaidy, L.~A. Al-Hbeti, Segmenting finger inner
  surface for the purpose of human recognition, in: 2019 2nd International
  Conference on Engineering Technology and its Applications (IICETA), IEEE,
  2019, pp. 105--110.

\bibitem{Bhaskar2014Hand}
B.~Bhaskar, S.~Veluchamy, Hand based multibiometric authentication using local
  feature extraction, in: International Conference on Recent Trends in
  Information Technology (ICRTIT), 2014, pp. 1--5.

\bibitem{Michael2010Robust}
G.~Michael, T.~Connie, A.~Jin, Robust palm print and knuckle print recognition
  system using a contactless approach, in: 5th IEEE Conference on Industrial
  Electronics and Applications (ICIEA), 2010, pp. 323--329.

\bibitem{Ribaric2005ABiometric}
S.~Ribaric, I.~Fratric, A biometric identification system based on eigenpalm
  and eigenfinger features, IEEE Transactions on Pattern Analysis and Machine
  Intelligence 27~(11) (2005) 1698--1709.

\bibitem{Al-Nima2017Robust}
R.~R.~O. Al-Nima, S.~S. Dlay, S.~A.~M. Al-Sumaidaee, W.~L. Woo, J.~A. Chambers,
  Robust feature extraction and salvage schemes for finger texture based
  biometrics, IET Biometrics 6~(2) (2017) 43--52.
\newblock \href {https://doi.org/10.1049/iet-bmt.2016.0090}
  {\path{doi:10.1049/iet-bmt.2016.0090}}.

\bibitem{Al-Nima2017finger}
R.~Al-Nima, M.~Abdullah, M.~Al-Kaltakchi, S.~Dlay, W.~Woo, J.~Chambers, Finger
  texture biometric verification exploiting multi-scale sobel angles local
  binary pattern features and score-based fusion, Elsevier, Digital Signal
  Processing 70 (2017) 178--189.

\bibitem{Al-Nima2017efficient}
R.~R. Al-Nima, S.~S. Dlay, W.~L. Woo, J.~A. Chambers, Efficient finger
  segmentation robust to hand alignment in imaging with application to human
  verification, in: 5th IEEE International Workshop on Biometrics and Forensics
  (IWBF), 2017, pp. 1--6.

\bibitem{Ribaric2005Anonline}
S.~Ribaric, I.~Fratric, An online biometric authentication system based on
  eigenfingers and finger-geometry, in: 13th European Signal Processing
  Conference, 2005, pp. 1--4.

\bibitem{Ferrer2007Low}
M.~Ferrer, A.~Morales, C.~Travieso, J.~Alonso, Low cost multimodal biometric
  identification system based on hand geometry, palm and finger print texture,
  in: 41st Annual IEEE International Carnahan Conference on Security
  Technology, 2007, pp. 52--58.

\bibitem{ying2007identity}
H.~Ying, T.~Tieniu, S.~Zhenan, H.~Yufei, Identity verification by using
  handprint, in: International Conference on Biometrics, Springer, 2007, pp.
  328--337.

\bibitem{Pavesic2009Finger-based}
N.~Pavesic, S.~Ribaric, B.~Grad, Finger-based personal authentication: a
  comparison of feature-extraction methods based on principal component
  analysis, most discriminant features and regularised-direct linear
  discriminant analysis, IET Signal Processing 3~(4) (2009) 269--281.

\bibitem{Al-Nima2015Human}
R.~R. Al-Nima, S.~S. Dlay, W.~L. Woo, J.~A. Chambers, Human authentication with
  finger textures based on image feature enhancement, in: 2nd IET International
  Conference on Intelligent Signal Processing (ISP), 2015.

\bibitem{michael2010innovative}
G.~K.~O. Michael, T.~Connie, A.~T.~B. Jin, An innovative contactless palm print
  and knuckle print recognition system, Pattern Recognition Letters 31~(12)
  (2010) 1708--1719.

\bibitem{Kanhangad2011AUnified}
V.~Kanhangad, A.~Kumar, D.~Zhang, A unified framework for contactless hand
  verification, IEEE Transactions on Information Forensics and Security 6~(3)
  (2011) 1014--1027.

\bibitem{Goh2010Bi-modal}
M.~K. Goh, C.~Tee, A.~B. Teoh, Bi-modal palm print and knuckle print
  recognition system, Journal of IT in Asia 3 (2010) 53--66.

\bibitem{kumar2011contactless}
A.~Kumar, Y.~Zhou, Contactless fingerprint identification using level zero
  features, in: IEEE Computer Society Conference on Computer Vision and Pattern
  Recognition Workshops (CVPRW), 2011, pp. 114--119.

\bibitem{Kumar2012Human}
{A. Kumar and Y. Zhou}, Human identification using finger images, IEEE
  Transactions on Image Processing 21~(4) (2012) 2228--2244.

\bibitem{zhang2012hand}
Y.~Zhang, D.~Sun, Z.~Qiu, Hand-based single sample biometrics recognition,
  Neural Computing and Applications 21~(8) (2012) 1835--1844.

\bibitem{otsu1975threshold}
N.~Otsu, A threshold selection method from gray-level histograms, IEEE
  Transactions on Systems, Man, and Cybernetics 9~(1) (1979) 62--66.

\bibitem{MAC2018contactless}
A.~{MAC}, K.~{Nguyen}, J.~{Banks}, V.~{Chandran}, Contactless multiple finger
  segments based identity verification using information fusion from higher
  order spectral invariants, in: 2018 15th IEEE International Conference on
  Advanced Video and Signal Based Surveillance (AVSS), 2018, pp. 1--6.
\newblock \href {https://doi.org/10.1109/AVSS.2018.8639153}
  {\path{doi:10.1109/AVSS.2018.8639153}}.

\bibitem{Jahan2018Contactless}
M.~{Akmal-Jahan}, J.~{Banks}, I.~{Tomeo-Reyes}, V.~{Chandran}, Contactless
  finger recognition using invariants from higher order spectra of ridge
  orientation profiles, in: 2018 25th IEEE International Conference on Image
  Processing (ICIP), 2018, pp. 2012--2016.
\newblock \href {https://doi.org/10.1109/ICIP.2018.8451664}
  {\path{doi:10.1109/ICIP.2018.8451664}}.

\bibitem{Zhang2010hand}
Y.~Zhang, D.~Sun, Z.~Qiu, Hand-based feature level fusion for single sample
  biometrics recognition, in: 2010 International Workshop on Emerging
  Techniques and Challenges for Hand-Based Biometrics, 2010, pp. 1--4.

\bibitem{Al-Nima2016ANovel}
R.~R.~O. Al-Nima, S.~S. Dlay, W.~L. Woo, J.~A. Chambers, A novel biometric
  approach to generate {ROC} curve from the probabilistic neural network, in:
  24th IEEE Signal Processing and Communication Application Conference (SIU),
  2016, pp. 141--144.

\bibitem{omar2018deep}
R.~R. Omar, T.~Han, S.~A.~M. Al-Sumaidaee, T.~Chen, Deep finger texture
  learning for verifying people, IET Biometrics 8 (2019) 40--48(8).
\newblock \href {https://doi.org/10.1049/iet-bmt.2018.5066}
  {\path{doi:10.1049/iet-bmt.2018.5066}}.

\bibitem{nanni2009multi}
L.~Nanni, A.~Lumini, A multi-matcher system based on knuckle-based features,
  Neural computing and applications 18~(1) (2009) 87--91.

\bibitem{al2018personal}
R.~Al-Nima, M.~Al-Kaltakchi, S.~Al-Sumaidaee, S.~Dlay, W.~Woo, T.~Han,
  J.~Chambers, Personal verification based on multi-spectral finger texture
  lighting images, IET Signal Processing.

\bibitem{al2018finger}
M.~T. Al-Kaltakchi, R.~R. Omar, H.~N. Abdullah, T.~Han, J.~A. Chambers, Finger
  texture verification systems based on multiple spectrum lighting sensors with
  four fusion levels, Iraqi Journal of Information \& Communications Technology
  1~(3) (2018) 1--16.

\bibitem{Kong2004Competitive}
A.-K. Kong, D.~Zhang, Competitive coding scheme for palmprint verification, in:
  Proceedings of the 17th International Conference on Pattern Recognition
  (ICPR), Vol.~1, 2004, pp. 520--523 Vol.1.

\bibitem{al2017multi}
S.~A. Al-Sumaidaee, M.~A. Abdullah, R.~R.~O. Al-Nima, S.~S. Dlay, J.~A.
  Chambers, Multi-gradient features and elongated quinary pattern encoding for
  image-based facial expression recognition, Pattern Recognition 71 (2017)
  249--263.

\bibitem{Petpon2009Face}
A.~Petpon, S.~Srisuk, Face recognition with local line binary pattern, in: 5th
  International Conference on Image and Graphics (ICIG), 2009, pp. 533--539.

\bibitem{young1998fundamentals}
I.~T. Young, J.~J. Gerbrands, L.~J. Van~Vliet, Fundamentals of {I}mage
  {P}rocessing, Delft University of Technology Delft, The Netherlands, 1998.

\bibitem{ISO2012Information}
{ISO}, {IEC}, Information technology — vocabulary — part 37: Biometrics,
  International Standard {ISO}/{IEC} 2382-37 (2012(E)) 1--21.

\bibitem{almahafzah2012multibiometric}
H.~AlMahafzah, M.~Imran, H.~Sheshadri, Multibiometric: Feature level fusion
  using {FKP} multi-instance biometric, IJCSI International Journal of Computer
  Science Issues 9~(3).

\bibitem{saugirouglu2009Intelligent}
{\c{S}}.~Sa{\u{g}}iro{\u{g}}lu, N.~{\"O}zkaya, An intelligent face features
  generation system from fingerprints, Turkish Journal of Electrical
  Engineering \& Computer Sciences 17~(2) (2009) 183--203.

\bibitem{meshoul2010novel}
S.~Meshoul, M.~Batouche, A novel approach for online signature verification
  using fisher based probabilistic neural network, in: IEEE Symposium on
  Computers and Communications (ISCC), IEEE, 2010, pp. 314--319.

\bibitem{Woods1997Generating}
K.~Woods, K.~Bowyer, Generating {ROC} curves for artificial neural networks,
  IEEE Transactions on Medical Imaging 16~(3) (1997) 329--337.

\bibitem{SAZAK2019TheMultiscale}
Çiğdem Sazak, C.~J. Nelson, B.~Obara, The multiscale bowler-hat transform for
  blood vessel enhancement in retinal images, Pattern Recognition 88 (2019) 739
  -- 750.
\newblock \href {https://doi.org/https://doi.org/10.1016/j.patcog.2018.10.011}
  {\path{doi:https://doi.org/10.1016/j.patcog.2018.10.011}}.

\bibitem{WILD2016Robust}
P.~Wild, P.~Radu, L.~Chen, J.~Ferryman, Robust multimodal face and fingerprint
  fusion in the presence of spoofing attacks, Pattern Recognition 50 (2016) 17
  -- 25.
\newblock \href {https://doi.org/https://doi.org/10.1016/j.patcog.2015.08.007}
  {\path{doi:https://doi.org/10.1016/j.patcog.2015.08.007}}.

\bibitem{jaswal2016knuckle}
G.~Jaswal, A.~Kaul, R.~Nath, Knuckle print biometrics and fusion
  schemes--overview, challenges, and solutions, ACM Computing Surveys (CSUR)
  49~(2) (2016) 34.

\end{thebibliography}
%\end{small}
\end{document}